\documentclass[authoryear, 11p, times]{elsarticle}
\usepackage{geometry}
\usepackage[authoryear]{natbib}
\usepackage{bm}
\usepackage{multirow}
\usepackage{amsmath, amssymb}
\DeclareMathAlphabet{\mathbbm}{U}{bbm}{m}{n}% from bbm.sty

\usepackage{mathtools}

\usepackage{microtype}
\usepackage{subfigure}
\usepackage[flushleft]{threeparttable}
\usepackage{tabularx,booktabs}
\RequirePackage[svgnames,dvipsnames]{xcolor}
\RequirePackage[colorlinks]{hyperref}
\colorlet{scolor}{black}
\colorlet{hscolor}{DarkSlateGrey}
\hypersetup{%
  pdfcreator={LaTeX3; cas-sc.cls; hyperref.sty},
  pdfproducer={pdfTeX;},
  linkcolor={hscolor},
  urlcolor={hscolor},
  citecolor={hscolor},
  filecolor={hscolor},
  menucolor={hscolor},
}

\begin{document}

\begin{frontmatter}
  \title{Traffic State Estimation from Vehicle Trajectories with Anisotropic Gaussian Processes}

  \author[1]{Fan Wu}
    \ead{fwu3@ualberta.ca}

  \author[2]{Zhanhong Cheng}
  \ead{zhanhong.cheng@mcgill.ca}

  \author[1]{Huiyu Chen}
  \ead{huiyu3@ualberta.ca}

  \author[1]{Zhijun Qiu \corref{cor1}}
  \ead{zqiu1@ualberta.ca}

  % Fifth author
  \author[2]{Lijun Sun}
  \ead{lijun.sun@mcgill.ca}

  \affiliation[1]{organization={Department of Civil and Environmental Engineering, University of Alberta},
  city={Edmonton},
  % citysep={}, % Uncomment if no comma needed between city and postcode
  postcode={T6G 1H9},
  state={Alberta},
  country={Canada}}

\affiliation[2]{organization={Department of Civil Engineering, McGill University},
  city={Montreal},
  postcode={H3A 0C3},
  state={Quebec},
  country={Canada}}

\cortext[cor1]{Corresponding author}

\begin{abstract}
Accurately monitoring {road traffic state} is crucial for various applications, including travel time prediction, traffic control, and traffic safety. However, the lack of sensors often results in incomplete traffic state data, making it challenging to obtain reliable information for decision-making. This paper proposes a novel method for imputing traffic state data using Gaussian processes (GP) to address this issue. We propose a kernel rotation re-parametrization scheme that transforms a standard isotropic GP kernel into an anisotropic kernel, which can better model the congestion propagation in traffic flow data. {The model parameters can be estimated by statistical inference using data from sparse probe vehicles or loop detectors.} Moreover, the rotated GP method provides statistical uncertainty quantification for the imputed traffic state, making it more reliable. We also extend our approach to a multi-output GP, which allows for simultaneously estimating the traffic state for multiple lanes. We evaluate our method using real-world traffic data from the Next Generation simulation (NGSIM) and HighD programs, {along with simulated data representing a traffic bottleneck scenario}. Considering current and future mixed traffic of connected vehicles (CVs) and human-driven vehicles (HVs), we experiment with the {traffic state estimation (TSE)} scheme from 5\% to 50\% available trajectories, mimicking different CV penetration rates in a mixed traffic environment. We also test the traffic state estimation when traffic flow information is obtained from loop detectors. {The results demonstrate the adaptability of our TSE method across different CV penetration rates and types of detectors, achieving state-of-the-art accuracy in scenarios with sparse observation rates.}
% \noindent Each keyword shall be separated by a \verb+\sep+ command.
\end{abstract}

% \includegraphics{figs/grabs.pdf}

% \begin{highlights}
% \item A new approach is proposed for {traffic state estimation (TSE)} using Gaussian process (GP) models with rotated anisotropic kernels that can capture the anisotropic correlation in traffic wave propagation. The rotation angle can be estimated from partially observed data and provides insight into the speed of congestion propagation in traffic waves.
% \item The proposed GP-based TSE method is a purely data-driven approach that does not require an external training dataset and provides statistical uncertainty quantification for the estimation, which is important for TSE under low connected vehicle (CV) penetration rates.
% \item The multi-output GP model is proposed for TSE on multiple lanes, which leverages the correlation between the traffic states of different lanes to improve TSE accuracy.
% \item Extensive experiments conducted on two real-world datasets {and a simulated dataset} showcase the accuracy and robustness of the proposed Gaussian Process-based TSE method under diverse CV penetration rates and detector types.
% \end{highlights}

% Keywords

\end{frontmatter}

\section{Introduction}
\label{sec: introduction}
Intelligent transportation systems (ITS) rely heavily on traffic state information, which is typically collected using a variety of detectors, such as loop detectors, video cameras, probe vehicles, and, more recently, connected vehicles (CVs). However, each type of detector has its limitations in terms of coverage and completeness of data. For instance, loop detectors are stationary sensors that only provide data at fixed locations, while video cameras require significant time and resources to process footage and must be installed on a high building or a gantry. As a result, these sensors are sparsely distributed in the traffic network, resulting in limited spatial coverage. In recent years, mobile sensors such as probe vehicles and CVs that can provide real-time traffic information, including speed and location, are playing an ever-important role in {traffic state estimation (TSE)}. However, because of the low penetration rate of CVs, the trajectories of CVs is sparse in both space and time. Therefore, an imputation method is needed to obtain the traffic state information in the entire spatiotemporal space, which would enable more accurate traffic control and management in ITS.

{Traffic state estimation refers to the inference of traffic state variables, such as density, speed, or other relevant variables, in a spatiotemporal domain by utilizing partially observed traffic data from detectors} \citep{seo2017traffic}. Generally, there are two types of TSE approaches: model-based and data-driven. Model-based TSE methods rely on traffic flow models and require strong prior knowledge, such as the capacity of the road, to accurately infer traffic state variables. Typical methods include first-order Lighthill-Whitham-Richards (LWR) model \citep{lighthill1955kinematic, richards1956shock} and high-order Payne-Whitham (PW) model \citep{payne1971model, whitham2011linear}. However, model-based TSE may not always be accurate because it may not fully capture the complexity of real-world traffic. Conversely, with massive traffic data and machine learning techniques available, TSE can be achieved in a purely data-driven manner, as demonstrated by some recent works \citep{wang2021low,thodi2022incorporating}. However, the training of a data-driven approach typically requires a large external training dataset and a validation dataset with full information. For example, many deep-learning-based TSE models \citep{thodi2022incorporating} are first trained on a traffic simulation dataset, and then applied to a real-world TSE problem. However, it may not always be possible to obtain an appropriate training dataset, and the external dataset may not be representative of the road segment with missing values. Therefore, there is a need to develop a data-driven TSE method without any external training dataset. For cases where the observed data is extremely sparse, we expect the TSE could also be able to provide statistical uncertainty quantification for the estimation.

To address the above research gap, we propose using Gaussian processes (GPs) \citep{rasmussen2006gaussian} for TSE. GPs are non-parametric Bayesian models that have been widely used for spatiotemporal kriging/imputation, providing a data-driven TSE approach that does not require an external training dataset. Additionally, GPs offer statistical uncertainty quantification for TSE. There are some studies that utilize the GPs in the calibration and evaluation of traffic flow models \citep{storm2022efficient, yuan2021macroscopic,cheng2022bayesian,liu2023gaussian, zhang2024bayesian}. However, conventional GP models are inadequate in modeling traffic flow data due to the non-stationarity and anisotropy caused by traffic wave propagation. Taking Fig.~\ref{fig: traffic speed} (a) as an example, the congestion wave propagates backward, generating directional spatiotemporal correlations that traditional GP kernels cannot model. To capture the anisotropic correlation in traffic wave propagation, we re-parameterized the GP kernel with a rotation angle. The kernel rotation angle indicates the speed of congestion propagation in traffic waves and can be estimated from partially observed data. We address the scalability issue of the GP model with variational sparse GP. Moreover, we propose using a multi-output GP model to simultaneously enable TSE on multiple lanes, rather than using several individual GPs. To test the TSE performance, we compare the proposed rotated GP with other imputation methods in NGSIM and highD datasets under different types and percentages of observed traffic information, which can also be regarded as CV penetration rates in the mixed traffic environment. { We also use simulated data to test the TSE performance under a traffic bottleneck scenario.} Experimental results demonstrate that the proposed rotated GP significantly outperforms other methods regarding accuracy and robustness for TSE under low CV penetration rates.

The contributions of this paper are summarized as follows:
\begin{itemize}[]
  \item A new approach is proposed for TSE using Gaussian process (GP) models with rotated anisotropic kernels that can capture the anisotropic correlation in traffic wave propagation. The rotation angle can be estimated from partially observed data, offering valuable insights into the speed of congestion propagation within traffic waves. Our approach elegantly merges statistical modeling with traffic flow theory.
  \item The proposed GP-based TSE method is a purely data-driven approach that does not require an external training dataset and provides statistical uncertainty quantification for the estimation, which is important for TSE under low CV penetration rates.
  \item  {The results from extensive experiments demonstrate the adaptability of our GP-based TSE method across different CV penetration rates and types of detectors, achieving state-of-the-art accuracy in scenarios with sparse observation rates.}
  \item The multi-output GP model is proposed for TSE on multiple lanes, which leverages the correlation between the traffic states of different lanes to improve TSE accuracy.
\end{itemize}

The remainder of the paper is organized as follows. We review related work on TSE in Section~\ref{sec: related work}. In Section~\ref{sec: methodology}, we describe the problem and the proposed method. The experimental settings and results are presented in Section~\ref{sec: case study}. Finally, we conclude the paper and discuss future research directions in Section~\ref{sec: conclusion}.

\section{Related work}
\label{sec: related work}
We can broadly classify existing traffic state estimation (TSE) models into model-based and data-driven methods, specifically model-based, data-driven, and streaming data-driven strategies \citep{seo2017traffic}. Model-based approaches adopt {macroscopic} traffic flow models mathematically to depict the traffic states; these models include first-order traffic models like the Lighthill-Whitham-Richards (LWR) model \citep{lighthill1955kinematic,richards1956shock}, high-order traffic models like Payne-Whitham (PW) model \citep{payne1971model, whitham2011linear} and the Aw-Rascle-Zhang (ARZ) model \citep{aw2000resurrection,zhang2002non, vishnoi2022traffic}, and their extensions. The model-based approach often performs the Data Assimilation (DA) or estimation with the traffic observation via the filter-based method. The most utilized one is the Kalman filter (KF) and its variants - KF-like techniques (e.g., the extended/unscented/ensembled Kalman filter) \citep{nanthawichit2003application, mihaylova2006unscented,wang2005real,work2010traffic, van2011localized, makridis2023adaptive}. Other methods are not oriented from the Kalman filter, such as particle filter (PF) \citep{mihaylova2007freeway}, adaptive smoothing filter (ASF) \citep{treiber2002reconstructing}, or others. A comprehensive review of the above methods can be found in \citep{seo2017traffic}. Although the model-based techniques can follow the traffic principles, these models rely a lot on the assumptions of traffic physics that can lead to numerical biases or approximation errors when the premises are not coherent with real-world data. In addition, model-based methods require substantial prior information on traffic dynamics.

Due to the availability of massive traffic data and the development of machine learning techniques, data-driven models get more attraction. Data-driven models usually utilize statistical or machine-learning approaches to infer traffic states from the spatiotemporal characteristics extracted from historical data (e.g., from sensors like loop detectors, cameras, or connected vehicles). For example, various research incorporates the spatiotemporal features into data-driven models utilizing the following techniques like the auto-regressive integrated moving average (ARIMA) \citep{zhong2004estimation}, Bayesian network (BN) \citep{ni2005markov}, Kernel regression (KR) \citep{yin2012imputing}, k-nearest neighbors (kNN) \citep{tak2016data}, convolutional neural networks (CNN) and deep neural networks (DNN) \citep{jia2016traffic, thodi2022incorporating, rempe2022estimation, han2021estimation, shi2021physics, shi2021physics2}, graph embedding generative adversarial network (GE-GAN) \citep{xu2020ge}, tensor decomposition \citep{wang2021low}, and principal component analysis \citep{li2013efficient} et al. Besides, streaming-data-driven approaches are regarded as more robust against uncertainties while requiring a large amount of streaming data to perform the prediction. Some research contributes to the streaming-data-driven models, such as \citep{seo2015traffic} and \citep{florin2016variant}. One of the advantages of data-driven models is that they require less prior info on traffic dynamics and can be more accurate than model-based methods. The other advantage is that they do not need rigid assumptions for traffic principles, which may cause unexplainable estimation. Meanwhile, the data-driven models often depend on a large amount of training data.

In terms of data sources. Most research discussed thus far utilized static data from fixed-location sensors like loop detectors or static data combined with mobile data obtained from probe vehicles \citep{nanthawichit2003application,seo2015probe,yuan2014network}. Recently, more and more researchers investigated the trajectory-based TSE methods from the data. For example, \citep{seo2015estimation, kyriacou2022bayesian, seo2015probe,wang2021low} utilized the trajectory data of probe vehicles or connected-automated vehicles (CAVs) to estimate the traffic state of freeways. Moreover, there is a trend that combines model-based and data-driven models to develop ``physics-informed'' machine learning models for TSE \citep{shi2021physics, usama2022physics, shi2021physics2}.
% However, such trajectory-based TSE methods also need the spacing information obtained from probe vehicles but are uncommon in other datasets. Therefore, it remains a noticeable problem that the above sensors can only provide one type of data, like only trajectory data, and how to perform the TSE based on the limited data source is a gap.

Traditional fixed sensors are expensive to install and maintain throughout the road. With the development of Connected Vehicle (CV) technology, there exists an opportunity that such moving sensors can provide data sources at a relatively cheaper cost all over the road. Therefore, we still have to deal with the scenario where data is sparse because not all cars are CVs to provide information. There is some research about the TSE using CV technology. \citet{chen2019traffic} estimated traffic states like flow, density, and speed by proposing an algorithm based on CV Basic Safety Message (BSM) data. This study tested the Kalman Filter and cell transmission algorithm in a simulator. { \citet{rempe2017phase} proposed a phase-based smoothing method to estimate traffic states from floating car trajectories; this method is consistent with the three-phase traffic theory \citep{kerner1999physics}.} \citet{fountoulakis2017highway} proposed microscopic simulation research of the TSE with mixed traffic (CVs and conventional vehicles) via CV and spot-sensor data. At the same time, \citet{bekiaris2017highway} developed a model-based TSE approach for per-lane density estimation and an on-ramp and off-ramp flow estimation in the presence of connected vehicles. There have also been works discussed online and offline TSE with CV data in a Bayesian model \citep{kyriacou2022bayesian}.

% The above research related to CV is the simulation-based or model-based approach. However, the constraints or environment in the above study are restrictive and ideal in the real-world scenario. Meanwhile, the spatiotemporal information didn't incorporate into the models well. Therefore, it is necessary to dig into data-driven methods considering the spatiotemporal info from real-world data.

As presented here, the most similar research to our work is the adaptive smoothing interpolation (ASM) by \cite{treiber2011reconstructing}. The authors used the anisotropic features of traffic waves and developed a smoothing method to estimate the traffic speed profile. { The interpolation of ASM is a weighted sum of a free-flow component and a congested component. \citet{schreiter2010two} proposed two fast implementations of ASM by efficient matrix operations and Fast Fourier Transform (FFT), bringing improvements in computation time by two orders of magnitude. The classical ASM lacks a well-defined method to determine the model parameters. \cite{yang2022generalized} reformulate ASM using matrix completion, which can estimate the weight parameter by the Alternating Direction Method of Multipliers (ADMM) algorithm. \cite{yang2023generalized} proposed a neural network model based on ASM, which can learn its parameters from sparse data of road sensors.} In our study, the proposed GP approach is a probabilistic model that can learn the parameters and uncertainties from the data. Besides, the proposed method can be applied to the TSE problem on a continuous space without defining grids.

\section{Methodology}
\label{sec: methodology}
\subsection{ Problem formulation} \label{sec: problem formulation}

We aim to estimate the traffic state (speed in this paper) of a highway segment over a period of time, using data collected from fixed or moving sensors such as loop detectors and CVs. For a single lane of the highway segment, we denote $s$ as the spatial coordinate on the segment, $t$ as the temporal coordinate, and $y(s,t)$ as the traffic speed at location $s$ and time $t$. In practice, $s$ and $t$ are usually defined as discrete values on an $S \times T$ spatiotemporal grid. However, for our purposes, we can consider a general continuous space with a subscript $i$ such that $y_i = y(\mathbf{x}_i)$, where $\mathbf{x}_i=[s_i,t_i]^\top$ is a vector representing the spatiotemporal coordinate.

Assume we can obtain the traffic speed $\mathbf{y}_o=\left\{y_i\right\}_{i=1}^{n} $ at a set of spatiotemporal locations $X_o=\left\{\mathbf{x}_i\right\}_{i=1}^n$ using loop detectors or probe vehicles, where $n$ is the number of observations. Our goal is to estimate the traffic speed distribution of {$\mathbf{y}_*=\left\{y_i\right\}_{i=n+1}^{n+u}$} at unknown spatiotemporal locations {$X_*=\left\{\mathbf{x}_i\right\}_{i=n+1}^{n+u}$} given the observed data $\left\{X_o, \mathbf{y}_o \right\}$, {where $u$ is the number of points/locations whose traffic state is unknown}. For a highway segment with multiple lanes, the problem becomes estimating the joint distribution $p\left(\mathbf{y}^1_*, \cdots, \mathbf{y}_*^L \right)$ from observations $\left\{X_o^1, \mathbf{y}_o^1, \cdots, X_o^L, \mathbf{y}_o^L \right\}$, where $L$ is the number of lanes.

\subsection{Gaussian process regression} \label{sec: gaussian process regression}
For a single lane of the highway segment, we assume the observed traffic state $y_i$ consists of a ground truth value $f_i$ and a noise term $\varepsilon_i$:
\begin{equation}\label{eq:first}
y_i = f_i + \varepsilon_i,
\end{equation}
where $\varepsilon$ is an independent and identically distributed (i.i.d.) Gaussian noise with zero mean and variance $\sigma_{\varepsilon}^2$.

Assume the ground truth traffic state $f$ is a function of the spatiotemporal coordinate $\mathbf{x}$.  We can impose a GP prior \citep{rasmussen2006gaussian} to the function $f(\mathbf{x}) \sim \mathcal{GP}\left(\mu, k\right)$. { A GP is a distribution over functions (or in other words, a distribution with an infinite number of random variables). The ground truth traffic state that occurred within the spatiotemporal range being examined can be considered as a sample drawn from the GP. With the GP prior, any finite collection of $\mathbf{f}\in \mathbb{R}^N$ at spatiotemporal location $X$ is assumed to follow a multivariate Gaussian distribution:
\begin{equation}
	\mathbf{f} = f(X) = \left[f(\mathbf{x}_1), \cdots, f(\mathbf{x}_N)\right]^\top \sim \mathcal{N}(\bm{\mu}, K),
\end{equation}
}
where the mean is often set to be zero $\bm{\mu}=\mathbf{0}$, and the covariance matrix $K$  is defined by a kernel function $k$ such that $K\left[i,j\right]=k(\mathbf{x}_i, \mathbf{x}_j)$. For example, the commonly used squared exponential (SE) kernel takes the form:
\begin{equation}\label{eq:se kernel}
	k_{\mathrm{SE}}(\mathbf{x}_i,\mathbf{x}_j) = \sigma^2 \exp \left( -\frac{1}{2\ell^2} \|\mathbf{x}_i-\mathbf{x}_j \|^2 \right),
\end{equation}
where the length scale $\ell$ determines how far apart two points in the input space can still be considered similar; the kernel variance $\sigma^2$ determines how far the function values can be from the mean. The kernel hyper-parameters and the noise variance $\bm{\theta}=\left\{ \sigma^2, \ell, \sigma_{\varepsilon}^2 \right\}$ can be estimated from the observed data using Maximum Marginal likelihood (MML) estimation.

Taking advantage of the conditional Gaussian distribution, the posterior distribution of traffic state $\mathbf{f}_*$ at the unknown spatiotemporal locations $\mathbf{X}_*$ given observed data can be obtained by:
\begin{align}
p\left(\mathbf{f}_* | X_*, X_o, \mathbf{y}_o\right) &\sim \mathcal{N}\left(\bar{\mathbf{f}}_*, \mathrm{cov}(\mathbf{f}_*) \right), \\
\bar{\mathbf{f}}_* &= K_{n*}^\top \left(K_{nn}+\sigma_{\varepsilon}^2 I\right)^{-1} \mathbf{y}_o, \\
\mathrm{cov}(\mathbf{f}_*) &= K_{**} - K_{n*}^\top \left(K_{nn}+\sigma_{\varepsilon}^2 I\right)^{-1} K_{n*},
\end{align}
where matrices $K_{nn}$, $K_{**}$, and $K_{n*}$ represent the kernel matrices evaluated at observed locations {(size $n\times n$)}, unknown locations {(size $u\times u$)}, and between observed and unknown locations {(size $n\times u$)}, respectively. Next, the distribution of $\mathbf{y}_*$ can be readily obtained by Eq.~\eqref{eq:first}.

\subsection{Rotated anisotropic kernel} \label{sec: anisotropic kernel}
Most GP kernels, such as the SE kernel in Eq.~\eqref{eq:se kernel}, are isotropic, meaning the covariance is only a function of $\|\mathbf{x}_i -\mathbf{x}_j\|$ and is invariant to the directions between $\mathbf{x}_i$ and $\mathbf{x}_j$. {The traffic wave, however, exhibits anisotropic behavior as it propagates along a specific direction.} Although an isotropic kernel can have anisotropic properties by using different length scales on different dimensions (i.e., implementing Automatic Relevance Determination (ARD) \citep{neal1996bayesian}), the ARD kernel is a very limited form and is still incapable of modeling the correlation propagates along a spatiotemporal direction.

\begin{figure}[!ht]
  \centering
  \includegraphics[width=0.4\textwidth]{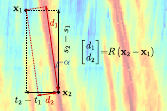}
  \caption{Illustration of the rotated coordinates.}
  \label{fig: rotate}
\end{figure}

Without loss of generality, let us consider the ``squared distance'' between $\mathbf{x}_i$ and $\mathbf{x}_j$ in an ARD kernel:
\begin{equation}
	d(\mathbf{x}_i, \mathbf{x}_j)^2 = (\mathbf{x}_i-\mathbf{x}_j)^{\top}M(\mathbf{x}_i-\mathbf{x}_j),
\end{equation}
where $M$ is a diagonal matrix with the $d$-th diagonal element being $\ell_d^{-2}$, specifying the dimension-specific length-scale. The diagonal structure of $M$ makes the length-scale along the spatial and temporal directions independent. To account for the traffic wave propagation, we introduce a rotation angle $\alpha$, a new hyper-parameter, which is the angle between the traffic wave and the space direction, as shown in Fig.~\ref{fig: rotate}. Then, we can measure the directional covariance using the following rotated squared distance: Then, we can measure the directional covariance using the following rotated squared distance:

\begin{equation}
\begin{split}
  d_{\mathrm{rot}}\left(\mathbf{x}_i, \mathbf{x}_j\right)^2 &=\left(R\left(\mathbf{x}_i-\mathbf{x}_j\right)\right)^{\top} M\left(R\left(\mathbf{x}_i-\mathbf{x}_j\right)\right) \\
&= (\mathbf{x}_i-\mathbf{x}_j)^{\top} \left(R^\top M R\right)(\mathbf{x}_i-\mathbf{x}_j), \label{eq:rotated_dist}\\
\end{split}
\end{equation}
\begin{align}
R &=\left[\begin{array}{cc}
\cos \alpha & -\sin \alpha \\
\sin \alpha & \cos \alpha
\end{array}\right].
\end{align}

% \begin{align}
% d_{\mathrm{rot}}\left(\mathbf{x}_i, \mathbf{x}_j\right)^2 &=\left(R\left(\mathbf{x}_i-\mathbf{x}_j\right)\right)^{\top} M\left(R\left(\mathbf{x}_i-\mathbf{x}_j\right)\right) \\
% &= (\mathbf{x}_i-\mathbf{x}_j)^{\top} \left(R^\top M R\right)(\mathbf{x}_i-\mathbf{x}_j), \label{eq:rotated_dist}\\
% R &=\left[\begin{array}{cc}
% \cos \alpha & -\sin \alpha \\
% \sin \alpha & \cos \alpha
% \end{array}\right] .
% \end{align}
The matrix $R$ is a rotation matrix. Eq.~\ref{eq:rotated_dist} can be used in general kernel functions. For example, the rotated squared distance can be used to define a rotated anisotropic SE kernel:
\begin{equation}
	k_{\mathrm{SE}}(\mathbf{x}_i,\mathbf{x}_j) = \sigma^2 \exp \left( -\frac{1}{2} d_{\mathrm{rot}}(\mathbf{x}_i,\mathbf{x}_j)^2 \right).
\end{equation}
The same transformation applies to other kernel functions, such as Mat\'ern kernels and the rational quadratic kernel.

A graphical illustration of our method is shown in Fig.~\ref{fig: rotate}. The intuition behind the proposed kernel is the rotation of the coordinates. The rotation angle $\alpha$ can be used to measure the speed of congestion propagation in the traffic wave. Similar to other hyper-parameters, the angle $\alpha$ can be estimated from the observed data.

\subsection{Model inference with variational sparse GP}\label{sec:sparse GP}
The computational complexity of an MML estimation of GP scales cubically with the number of data points, which limits its applicability to large datasets. Therefore, we use the variational sparse GP (VSGP) \citep{titsias2009variational} for scalable inference. VSGP introduces a set of $m$ inducing points at $Z=\left\{ \mathbf{z}_1, \cdots, \mathbf{z}_m \right\}$ that act as a sparse approximation to the full Gaussian process. The model assumes that the function values at the inducing points follow the same GP prior, and the posterior distribution of the function values is approximated by a Gaussian distribution conditioned on the inducing variables. The locations of inducing points can be optimized as other hyper-parameters.

The model parameters, including hyper-parameters $\bm{\theta}$ and the locations of inducing variables $Z$, are learned by maximizing the evidence lower bound (ELBO), which is a lower bound of the log marginal likelihood of the observed data. The ELBO of VSGP derived by Titsias \citep{titsias2009variational} is:
\begin{equation}\label{eq:elbo}
\log p(\mathbf{y}_o) \geq \log \mathcal{N}\left(\mathbf{y}_o | \mathbf{0}, Q_{nn}+\sigma_{\varepsilon}^2 I\right) - \frac{1}{2\sigma_{\varepsilon}^2} \operatorname{tr}\left(K_{nn} - Q_{nn}\right),
\end{equation}
where $Q_{nn}=K_{nm}K_{mm}^{-1}K_{nm}^\top$, matrices $K_{mm}$ and $K_{nm}$ are the kernel evaluated at the inducing points, and between the observed locations and inducing points, respectively. We can interpret the ELBO as the sum of the approximate log marginal likelihood and a regularization term $- \frac{1}{2\sigma_{\varepsilon}^2} \operatorname{tr}\left(K_{nn} - Q_{nn}\right)$. The regularization term minimizes the squared error of predicting the training latent function values $\mathbf{f}_n$ from the inducing variables. Eq.~\eqref{eq:elbo} can be simplified with the Woodbury matrix identity, and the time complexity of VSGP is $O(mn^2)$.

The posterior distribution of function values at unknown location $\mathbf{x}_*$ is given by the integral $p(\mathbf{f}_*) = \int p(\mathbf{f}_*|\mathbf{f}_z) p(\mathbf{f}_z) d\mathbf{f}_z$, which is a Gaussian distribution with the mean and covariance:
\begin{align}
\bar{\mathbf{f}}_* &= K_{*m} K_{mm}^{-1} \bar{\mathbf{f}}_z, \\
\operatorname{cov}(\mathbf{f}_*) &= K_{**} - K_{m*}^\top K_{mm}^{-1} K_{m*} + K_{m*}^\top K_{mm}^{-1} \Lambda K_{mm}^{-1} K_{m*},\label{eq:cov_svgp}
\end{align}
where $\bar{\mathbf{f}}_z=\sigma_{\varepsilon}^{-2}\Lambda^{-1}K_{mm}^{-1}K_{mn}\mathbf{y}_o$ is the posterior mean of the inducing variables, and $\Lambda=K_{mm}^{-1}+ \sigma_{\varepsilon}^{-2}K_{mm}^{-1}K_{mn}K_{nm}K_{mm}^{-1}$ is the posterior precision matrix of the inducing variables.

% The VSGP model has been shown to achieve competitive performance with standard Gaussian process regression while reducing the computational complexity from cubic to linear or sublinear, depending on the number of inducing variables. It has been applied in various domains, including computer vision, robotics, and neuroscience, to model complex data with large numbers of observations.

% The variational sparse GP model \cite{titsias2009variational} is a scalable GP model that can be used to estimate the hyper-parameters. The variational sparse GP model is a GP model with a sparse set of inducing points $\mathbf{Z}=\left\{ \mathbf{z}_1, \cdots, \mathbf{z}_M \right\}$, where $M$ is the number of inducing points. The sparse GP model is defined as:

\subsection{Multi-output GP}
\label{sec: multi-output GP}
{ The traffic states of neighboring lanes of the same road are highly correlated. Because drivers can choose a less congested lane to travel and thus the traffic state of the neighboring lane reaches a similar condition.} The TSE in a highway segment with multiple lanes can be naturally modeled using a multi-output GP model \citep{wackernagel2003multivariate, bonilla2007multi}, also known as a coregionalized GP or co-kriging. Unlike using independent GP models for each lane, a multi-output GP model can leverage the correlation between the traffic states of different lanes to improve estimation accuracy. In Section~\ref{sec: case multi-lane}, we will demonstrate that the multi-output GP model can estimate traffic speed during a long period that has no observations in a lane, by utilizing information from the other lane.

The probe vehicles from different lanes locate at different spatiotemporal locations, which is referred to as heterotopic data in the multi-output GP literature. We model the traffic states of $L$ different lanes as a multi-output function $\bm{f}\left(\mathbf{x}\right)=\left[f^1(\mathbf{x}),\cdots, f^L(\mathbf{x})\right]^\top$ with a GP prior. The covariance of the $i$-th output at $\mathbf{x}$ and the $j$-th output at $\mathbf{x}'$ is given by the kernel function:
\begin{equation}
k_{\operatorname{multi}}\left(f^i(\mathbf{x}), f^j(\mathbf{x}')\right) = k(\mathbf{x}, \mathbf{x}') B\left[i,j\right],
\end{equation}
where $B$ is an $L\times L$ symmetric and positive-definite matrix parametrized by $B=AA^\top$, and $A\in\mathbb{R}^{L\times r}$ is a parameter to learn, $r$ is the rank of $A$. This kernel parametrization is also called the intrinsic model of coregionalization \citep{wackernagel2003multivariate} in the geostatistics literature. One can view the multi-output kernel $k_{\operatorname{multi}}$ as functions on an extended input space with the index of the lane, which allows for using the same inference procedure as the single-output GP model.

\section{Case Study}
\label{sec: case study}
% In this section, we evaluate the proposed GP-based TSE method on two real-world datasets: the NGSIM \citep{simulation2007us} traffic trajectory data and the HighD \citep{highDdataset} naturalistic vehicle trajectory data. Then, considering current and future mixed traffic of CVs and HVs, we compare TSE performance under different CV penetration rates and compare the proposed rotated GP method with other benchmark models, including adaptive smoothing interpolation method (ASM) \citep{treiber2011reconstructing}, Spatiotemporal Hankel Low-Rank Tensor Completion (STH-LRTC) \citep{wang2021low}, and standard GP with ARD kernel. We also explore the use of multi-output rotated GP for TSE on multiple lanes.

We evaluate the proposed GP-based TSE method on two real-world datasets: the NGSIM \citep{simulation2007us} traffic trajectory data and the HighD \citep{highDdataset} naturalistic vehicle trajectory data. In Section~\ref{sec: overall performance}, we compare the TSE performance of the proposed model with a set of benchmark models under different CV penetration rates. In Section~\ref{sec: loop performance}, we evaluate the TSE performance when assuming using loop detector data. The uncertainty quantification and computational time of the proposed method are further analyzed in Section~\ref{sec: uncertainty} and \ref{sec: computation}. Finally, We also explore the use of multi-output rotated GP for TSE on multiple lanes. The code and data associated with this paper are available at \url{https://github.com/Lucky-Fan/GP_TSE}.

% In this section, we apply the proposed GP methods for TSE on two real-world datasets and compare them with other benchmark models to evaluate performances. We begin by introducing the data and experiment settings. One of the datasets is the traffic trajectory data from the Next Generation Simulation (NGSIM) program; the other is the naturalistic vehicle trajectories from HighD Dataset \cite{highDdataset}. Then, considering current and future mixed traffic of CVs and HVs, we compare the TSE under CV penetration rates of 5\%, 10\%, 20\%, 30\%, 40\%, and 50\%. Next, we compare the proposed rotated GP method with the adaptive smoothing interpolation method (ASM) \cite{treiber2011reconstructing}, Spatiotemporal Hankel Low-Rank Tensor Completion (STH-LRTC) \cite{wang2021low}, and standard GP with ARD kernel. Finally, we also test using multi-output rotated GP to estimate the traffic state of multiple lanes simultaneously.

% which means assuming a certain proportion of CVs as probe vehicles as a training set to imitate the penetration rates of connected vehicles of mixed traffic.
% which integrates various traffic states into the spatiotemporal kernel model. Meanwhile, we also compare the proposed rotated GP method with the
 % Furthermore, we compare the proposed rotated GP model with GP with standard ARD SE or Mat\'ern kernels (GP-ARD) to show the advantages over traditional GP models.

\subsection{Data and experimental setup} \label{sec: experimental setup}
We test the proposed rotated GP TSE using the trajectories from two real-world datasets, namely NGSIM \citep{simulation2007us} and HighD \citep{highDdataset}. Both datasets provide detailed information about each vehicle's trajectory, such as vehicle ID, recording frame, time, location, velocity, lane, etc. This allows us to use the ground truth traffic state to evaluate the accuracy of TSE. In the case of NGSIM, we focus on the traffic data from lane 2 of US Highway 101. For HighD, we utilize data from two lanes of a German highway. { These lanes are selected as they display some very representative stop-and-go traffic waves.} The specific details of the two datasets are provided below:

\begin{itemize}[]
\item The NGSIM data: We use vehicle trajectories extracted from video cameras on lane 2 of US highway 101. In contrast to the previous work by Wang et al. \citep{wang2021low}, our experiment covers a longer road segment of 600 meters and a larger time range of 2500 seconds. We extract the complete data and focus on the traffic state at a $200\times 500$ spatiotemporal grid with a resolution of 3 meters and 5 seconds, where the traffic state is defined as the average vehicle speed in each grid cell. Fig.~\ref{fig: traffic speed} (a) and (b) show the traffic speed maps of the entire dataset and samples of observed trajectories under a 5\% penetration rate, respectively.

\item The HighD data: This dataset provides naturalistic vehicle trajectories recorded on German highways using drones. The dataset includes 60 recordings from six different locations; each recording is identified by track ID. In our study, we focus on the recording with track ID 25. Where the full drive length of vehicles during the road segment is 1120346.1 meters, and the time range is 80676.08 seconds. To make the most of the data, we extract traffic state in a spatiotemporal grid of size $100\times 220$ with a resolution of 4 meters and 5 seconds, representing a domain of 400 meters and 1100 seconds. The average vehicle speed is calculated to describe the traffic state at each cell. We use the data from lane 4 for the TSE of a single lane in Table.~\ref{tab: detailed HighD}, and we use the data from lane 3 and lane 4 to test using multi-output GP for TSE.
\end{itemize}

When using CVs as probe vehicles, we set 5\%, 10\%, 20\%, 30\%, 40\%, and 50\% as the penetration rate of CVs and assume only the trajectories of CVs are observed (i.e., the training data). Under each CV penetration rate, we repeat the experiment 10 times with different random draws of trajectories. Note that we define spatiotemporal grids to make an easy comparison with other models, although GP can make TSE on a continuous space without defining grids. Overall, the NGSIM and HighD datasets provide rich sources of data for evaluating the effectiveness and efficiency of our approach and baselines under different scenarios.

% the vehicles as probe vehicles and used their trajectories as a training set to estimate the traffic state at the unknown locations. We can also assume the probe vehicles are connected vehicles, and 5\% to 50\% can be regarded as CVs' penetration rate (PR). Under each penetration rate, we run the experiment 10 times using randomly drawn trajectories to test the accuracy and robustness of different TSE algorithms.
% We also randomly extract 5\%, 10\%, 20\%, 30\%, 40\%, and 50\% of the vehicle trajectories from the complete data as observed sets. The experiments were run 10 times with randomly sampled vehicles under each penetration rate. We examine our rotated GP and benchmark models on the trajectory data of lane 4.

\subsection{Baseline models and hyper-parameters} \label{sec: benchmark}
In the following, we refer to the proposed GP based on rotated kernels as ``GP-rotated''. We use the following baselines to compare the performance of GP-rotated with other methods:
\begin{itemize}
    \item The adaptive smoothing interpolation method (ASM) \citep{treiber2011reconstructing}: It is an interpolation method for estimating traffic states. This method resembles the proposed rotated GP in terms of considering the traffic wave propagation using an anisotropic interpolation. { We set the propagation speed of congestion traffic to be -19.87 km/h and -17.86 km/h for the NGSIM and HighD datasets, respectively, which is based on the estimated hyper-parameters of the GP-rotated method (see discussion in Section~\ref{sec: overall performance}). The space smoothing width and time smoothing width are set as 200 m and 10 s, respectively, based on an initial setting suggested by \citep{treiber2011reconstructing} and a quick validation. Other parameters adopt the settings of \citet{treiber2011reconstructing}.}
     \item Spatiotemporal Hankel Low-Rank Tensor Completion (STH-LRTC) \citep{wang2021low}: It transforms the original speed matrix into a tensor using spatial and temporal delay embedding. Then, the approach estimated the traffic state matrix by conducting inverse Hankelization on the delay-embedding tensor imputed by a low-rank model. The key parameters include the embedding lengths $\tau_s$ and $\tau_t$. We adopt the same hyper-parameter settings $\tau_s=40$ and $\tau_t=30$ as the author. But we do find the method produces poor results in certain cases with extremely low CV penetration rates. Therefore, we increase the $\tau_s$ and $\tau_t$ with case-specific tuning, as noted in Table~\ref{tab: computation}.
    \item Gaussian process regression with standard ARD kernels (GP-ARD): It extends the basic GP by allowing the kernel function to have a separate length scale parameter for each input dimension, which enables the model to automatically determine the importance of each input variable in predicting the output variable. The hyper-parameters are learned from data using the VSGP approach.
\end{itemize}

 We use Mat\'ern kernel \citep{rasmussen2006gaussian} as the basic kernel function for all GP models. We set the number of inducing points $m=\min \left(0.02 n, 500\right)$, and the initial locations of inducing points are randomly distributed on the grid. As introduced in Section~\ref{sec:sparse GP}, the hyper-parameters of GP-rotated can be learned from the observed data, which could be time-consuming for a large dataset. However, one may not need to repeatedly learn the hyper-parameters for the same highway segment in reality. Therefore, we also test the performance of the proposed GP using pre-trained hyper-parameters, referred to as "P-GP-rotated". Please note that we do not include comparisons with deep-learning-based models as many of them depend on external training datasets and are not available as open-source.

It's worth noting that the training data (obtained from CVs) and test data (obtained from all vehicles) for a cell with CV trajectories may differ since the speed may be calculated from different numbers of vehicles in the training and test data. {TSE values are only used for the cells without any CV trajectories. For cells with observed trajectories, we directly use the speed from the training data, because we found that observed values on these cells are closer to the test data than the estimated value, no matter what TSE method is used. This can be attributed to the utilization of high-quality datasets.} Finally, we use the root mean squared error (RMSE) and mean absolute error (MAE) as shown in the following to evaluate the performance of different TSE models:
\begin{align}
\text{RMSE} &= \sqrt{\frac{\sum_{l}\sum_{s}\sum_{t} (y^l(s,t)-\hat{y}^l(s,t))^2}{S T L}}, \\
\text{MAE} &= \frac{\sum_{l}\sum_{s}\sum_{t} |y^l(s,t)-\hat{y}^l(s,t)|}{S T L}.
\end{align}

% To assess the performance of different models, we utilize the root mean squared error (RMSE) and the mean absolute error (MAE). hyper-parameters are also crucial in determining model performance. We use the hyper-parameters from \citet{treiber2011reconstructing} for the ASM and those from \citet{wang2021low} for the STH-LRTC. The hyper-parameters of GP-ARD and GP-rotated are estimated from the training set using the maximum likelihood method.

% We applied ASM, STH-LRTC, and GP-ARD to the two datasets mentioned above and found that these methods could not outperform the proposed rotated GP model. Furthermore, our proposed method has a faster computation compared to these existing approaches, which is shown in Section \ref{sec: computation}. Additionally, the rotated GP method can be extended to co-regionalized GP to perform TSE on multiple lanes simultaneously, which is demonstrated in Section \ref{sec: case multi-lane}.

% Finally, we utilize the root mean squared error (RMSE) and the mean absolute error (MAE).
% hyper-parameters are also crucial in determining model performance. We use the hyper-parameters from \citet{treiber2011reconstructing} for the ASM and those from \citet{wang2021low} for the STH-LRTC. The hyper-parameters of GP-ARD and GP-rotated are estimated from the training set using the maximum likelihood method.
% \subsection{ Case of NGSIM dataset} \label{sec: performance evaluation}

\subsection{TSE from vehicle trajectories} \label{sec: overall performance}
\begin{figure*}[!ht]
  \begin{center}
  \includegraphics[width=0.8\textwidth]{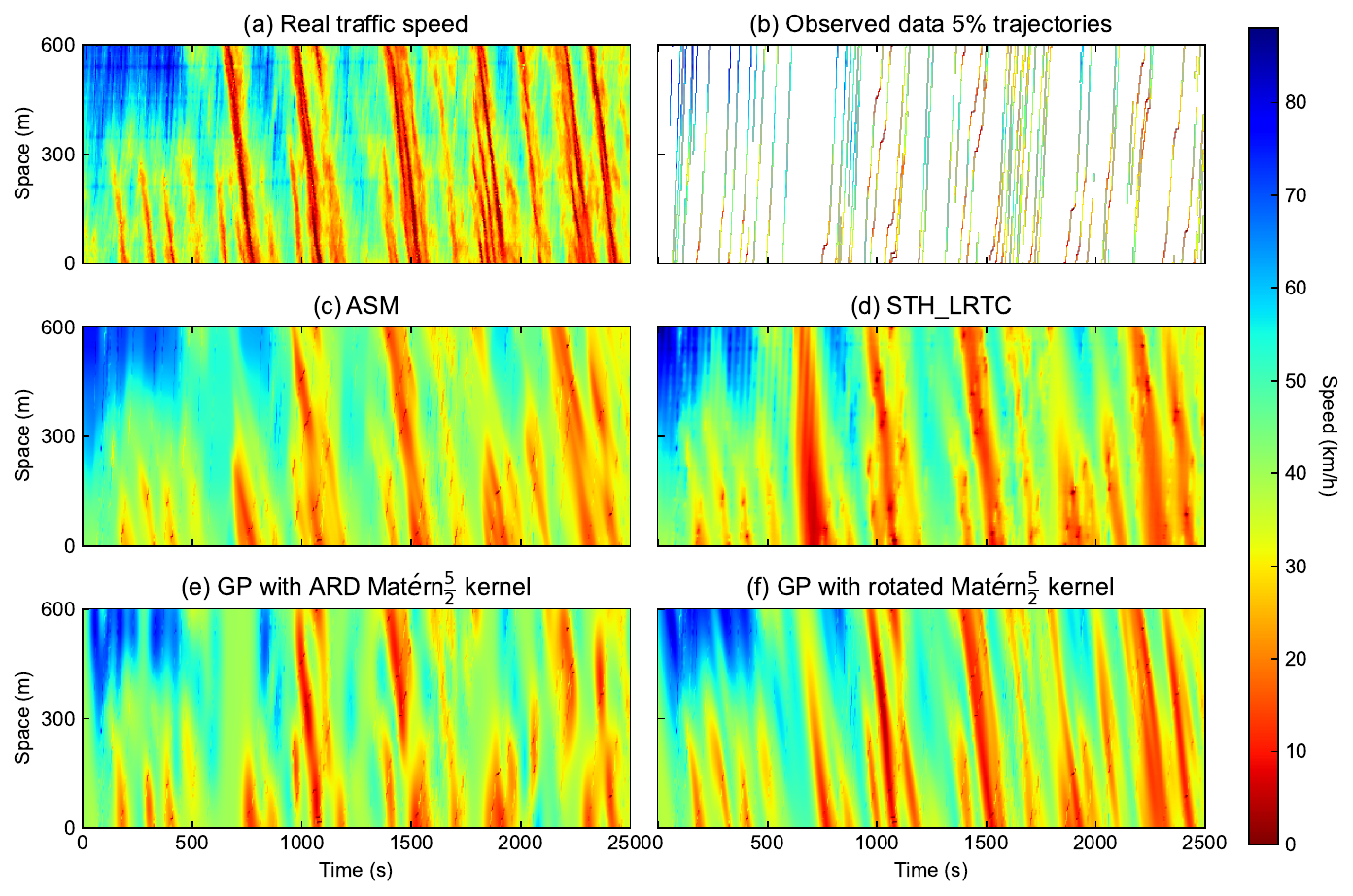}
  \caption{A TSE experiment on the NGSIM dataset with 5\% CVs penetration rate. The observed trajectories are superimposed on the TSE results. (a) The traffic speed of the full dataset. (b) The traffic speed of observed trajectories. (c) The traffic speed estimated by the ASM method. (d) The traffic speed estimated by the STH-LRTC method. (e) The traffic speed estimated by the GP with ARD Mat\'ern$\frac{5}{2}$ kernel. (f) The traffic speed estimated by the GP with the proposed rotated Mat\'ern$\frac{5}{2}$ kernel.}
  \label{fig: traffic speed}
  \end{center}
\end{figure*}

We begin by visually examining the TSE performance of different methods under a 5\% CV penetration rate using the NGSIM dataset. Fig.~\ref{fig: traffic speed} displays the results. Fig.~\ref{fig: traffic speed} (a) shows the ground truth traffic speed map of all trajectories, exhibiting complex traffic dynamics evolution with shockwaves, making it a suitable dataset for experimentation. Fig.~\ref{fig: traffic speed} (b) displays one of the randomly selected 5\% training datasets from ten independent experiments.

% because we varied the training data for each of the ten experiment repetitions.
% We randomly selected 5\% of trajectories from the NGSIM dataset to estimate the complete traffic speed profile. and Table~\ref{tab: detailed NGSIM}.

By comparing Fig.~\ref{fig: traffic speed} (e) and (f), we can observe that the proposed GP-rotated captures the directional traffic speed correlations that traditional GP-ARD cannot model. When comparing the ASM in Fig.~\ref{fig: traffic speed} (c) with the proposed GP-rotated, we can find they both capture the congestion propagation in the traffic wave because they both use the idea of anisotropic kernels. However, the congestion speed estimated by the ASM is generally lower than the ground truth speed, which is caused by the ``smoothing'' operation in the ASM. The STH-LRTC in Fig.~\ref{fig: traffic speed} (d) also captures the congestion propagation in the traffic wave, but it performs poorly when there is a long period without CV trajectories (e.g., 550s-750s).

% is the second-best performing model, but we can observe some unnatural knots in the imputation results (Fig.~\ref{fig: traffic speed} (d)).

% Table~\ref{tab: detailed NGSIM} displays the average MAE (m/s) and RMSE (m/s) obtained from the ASM, STH-LRTC, GP-ARD, and GP-rotated methods. The proposed method outperforms the other scenarios concerning both the RMSE and the MAE. Additionally, the results reveal that the rotated Mat\'ern$\frac{5}{2}$ kernel performs better than the rotated SE kernel, perhaps because the Mat\'ern$\frac{5}{2}$ is less smooth than the SE kernel, making it more suitable for traffic speed estimation problems.

Next, we perform more extensive experiments to quantify the performance of the proposed method and the baselines under different penetration rates. For each penetration rate (percentage of trajectories). For each penetration rate (5\%, 10\%, 20\%, 30\%, 40\%, and 50\%), we repeat the experiments ten times with randomly selected vehicle trajectories from the complete dataset as the training set (see Section~\ref{sec: experimental setup}). The experiments were conducted on the NGSIM and HighD datasets. Table~\ref{tab: detailed NGSIM} and Table~\ref{tab: detailed HighD} display the average MAE (m/s) and RMSE (m/s) with standard deviation for each method under various penetration rates and datasets.

\begin{table*}[!htbp]
\scriptsize
  \centering
  \caption{TSE accuracy for the NGSIM dataset under different penetration rates: mean (std).}
  \setlength{\tabcolsep}{1pt}{
    \begin{tabularx}{\textwidth}{XXXXXXXXXXX}
    \toprule
    Method & \multicolumn{2}{c}{ASM} & \multicolumn{2}{c}{STH-LRTC} & \multicolumn{2}{c}{GP-ARD} & \multicolumn{2}{c}{GP-rotated} & \multicolumn{2}{c}{P-GP-rotated} \\
    \midrule
    Rate  & MAE   & RMSE  & MAE   & RMSE  & MAE   & RMSE  & MAE   & RMSE  & MAE   & RMSE \\
    \midrule
    0.05   & {5.11 (0.32)} & {7.04 (0.54)} & 5.51 (1.36) & 7.94 (2.38) & 6.02 (0.36) & 8.62 (0.56) & \textbf{4.85 (0.31)} & 6.74 (0.56) & 4.97 (0.29) & \textbf{6.74 (0.47)} \\
    0.1  & {4.09 (0.15)} & {5.72 (0.26)} & 4.19 (1.39) & 7.43 (5.62) & 4.35 (0.30) & 6.42 (0.58) & 3.82 (0.22) & 5.44 (0.47) & \textbf{3.79 (0.13)} & \textbf{5.19 (0.22)} \\
    0.2  & {3.28 (0.10)} & {4.82 (0.14)} & 3.01 (1.26) & 6.16 (7.14) & 3.07 (0.14) & 4.61 (0.26) & \textbf{2.81 (0.10)} & \textbf{4.10 (0.19)} & 2.98 (0.08) & 4.28 (0.12) \\
    0.3  & {2.73 (0.06)} & {4.26 (0.09)} & \textbf{2.09 (0.05)} & \textbf{3.17 (0.12)} & 2.43 (0.06) & 3.77 (0.11) & 2.27 (0.05) & 3.43 (0.10) & 2.48 (0.05) & 3.75 (0.08) \\
    0.4  & {2.29 (0.06)} & {3.83 (0.09)} & \textbf{1.75 (0.05)} & \textbf{2.81 (0.12)} & 2.03 (0.06) & 3.35 (0.11) & 1.92 (0.05) & 3.08 (0.09) & 2.09 (0.05) & 3.37 (0.09)  \\
    0.5  & {1.87 (0.05)} & {3.41 (0.09)} & \textbf{1.43 (0.04)} & \textbf{2.46 (0.11)} & 1.67 (0.04) & 2.96 (0.10) & 1.58 (0.04) & 2.71 (0.09) & 1.71 (0.04) & 2.98 (0.07) \\
    \bottomrule
    \end{tabularx}
    }%
  \label{tab: detailed NGSIM}%
\end{table*}%

\begin{table*}[!htbp]
\scriptsize
  \centering
  \caption{TSE accuracy for the HighD dataset under different penetration rates: mean (std).}
  \setlength{\tabcolsep}{1pt}{
    \begin{tabularx}{\textwidth}{XXXXXXXXXXX}
    \toprule
    Method & \multicolumn{2}{c}{ASM} & \multicolumn{2}{c}{STH-LRTC} & \multicolumn{2}{c}{GP-ARD} & \multicolumn{2}{c}{GP-rotated} & \multicolumn{2}{c}{P-GP-rotated} \\
    \midrule
    Rate  & MAE   & RMSE  & MAE   & RMSE  & MAE   & RMSE  & MAE   & RMSE  & MAE   & RMSE \\
    \midrule
    0.05   & {4.44 (0.31)} & {\textbf{6.03 (0.50)}} & 55.9 (29.3) & 121.5 (51.1) & 5.18 (1.59) & 7.19 (2.04) & 4.48 (0.50) & 6.27 (0.94) & \textbf{4.43 (0.27)} & 6.06 (0.43) \\
    0.1  & {3.32 (0.14)} & {4.67 (0.22)} & 3.19 (0.10) & 4.49 (0.21) & 3.22 (0.17) & 4.55 (0.28) & \textbf{3.18 (0.16)} & \textbf{4.45 (0.25)} & 3.55 (0.18) & 5.00 (0.30) \\
    0.2  & {2.55 (0.08)} & {3.79 (0.11)} & \textbf{2.15 (0.11)} & \textbf{3.12 (0.18)} & 2.23 (0.09) & 3.26 (0.15) & 2.23 (0.09) & 3.25 (0.13) & 2.43 (0.07) & 3.52 (0.11) \\
    0.3  & {2.08 (0.03)} & {3.30 (0.05)} & \textbf{1.65 (0.04)} & \textbf{2.49 (0.06)} & 1.71 (0.05) & 2.62 (0.10) & 1.71 (0.05) & 2.61 (0.10) & 1.89 (0.05) & 2.90 (0.09) \\
    0.4  & {1.68 (0.04)} & {2.85 (0.07)} & \textbf{1.31 (0.04)} & \textbf{2.07 (0.07)} & 1.36 (0.04) & 2.14 (0.09) & 1.35 (0.03) & 2.14 (0.08) & 1.49 (0.03) & 2.39 (0.06) \\
    0.5  & {1.37 (0.05)} & {2.49 (0.10)} & \textbf{1.05 (0.03)} & \textbf{1.75 (0.07)} & 1.09 (0.03) & 1.80 (0.07) & 1.09 (0.03) & 1.80 (0.06) & 1.19 (0.03) & 2.03 (0.05) \\
    \bottomrule
    \end{tabularx}%
    }
  \label{tab: detailed HighD}%
\end{table*}%

Table~\ref{tab: detailed NGSIM} and Table~\ref{tab: detailed HighD} illustrate that the performance of the STH-LRTC method is the best when the proportion of observed trajectories (penetration rate) is over 30\%. This is because the GP-based methods cannot capture the fine-grained texture in traffic flow, which is shown in Section~\ref{sec: uncertainty}. However, for the cases with sparse observations (CV penetration from 5\% to around 20\%), our proposed GP-rotated performs the best on both NGSIM and HighD datasets. As the decrease of CV penetration rate, the STH-LRTC may fail due to a large block of missing information, which can be observed in the HighD data under the 5\% rate. The corresponding MAE and RMSE values can be as high as 55.9 m/s and 121.5 m/s, respectively. The STH-LRTC may be unstable under low penetration rates, with high standard deviation in MAE and RMSE values. In contrast, our proposed GP method provides a very robust estimation, regardless of the percentage of probe vehicles. In the early stages of a mixed traffic environment with a low CV penetration rate, such as 5\% to 20\%, our proposed method is a suitable choice.

Our proposed method consistently outperforms the ASM benchmark. An advantage of ASM is that it considers the traffic propagation of both congestion and free flow. Therefore, ASM could produce a more natural traffic flow pattern, as demonstrated in the high-speed region (top left corner) of Fig. \ref{fig: traffic speed} (c). However, the ASM is not as good as the GP-rotated in estimating small shockwaves during the first 500 seconds (bottom left corner). { The proposed GP-rotated outperforms ASM in most cases (except for the highD dataset with 5\% penetration rate). The gaps between ASM and GP-rotated increase with a larger CV penetration rate.}

We also compare the GP-ARD and GP-rotated methods using the Mat\'ern$\frac{5}{2}$ kernel in Tables~\ref{tab: detailed NGSIM} and~\ref{tab: detailed HighD}. Our results show that the GP-rotated method consistently outperforms the GP-ARD method in terms of average MAE and RMSE, regardless of whether we use the pre-trained parameters or not. This can be attributed to the fact that the ARD kernel is isotropic, meaning it uses distance to measure covariance. However, traffic waves are anisotropic and propagate in spatiotemporal directions. Although GP-ARD can apply different length scales to different dimensions, it still fails to capture the directional covariance of traffic waves. On the other hand, our proposed anisotropic GP-rotated method with a directional hyper-parameter $\alpha$ accurately models the directional covariance of traffic waves. Furthermore, we observed that the numerical differences in MAE and RMSE between GP-ARD and GP-rotated decrease with an increase in the observation rate. This is because, with more observed trajectories, the GP-ARD method can utilize more information to overcome the directional limitation based on length scale adjustments.

We observe that the P-GP-rotated method can also achieve satisfactory TSE results and even outperform the GP-rotated in some low-penetration scenarios, possibly due to the overfitting of GP-rotated when the number of observed trajectories is small. However, as the CV penetration rate increases, the performance of P-GP-rotated is not as good as GP-rotated, likely because the inducing point locations in P-GP-rotated are not optimized.

\begin{figure}
    \centering
    \includegraphics[width=0.5\textwidth]{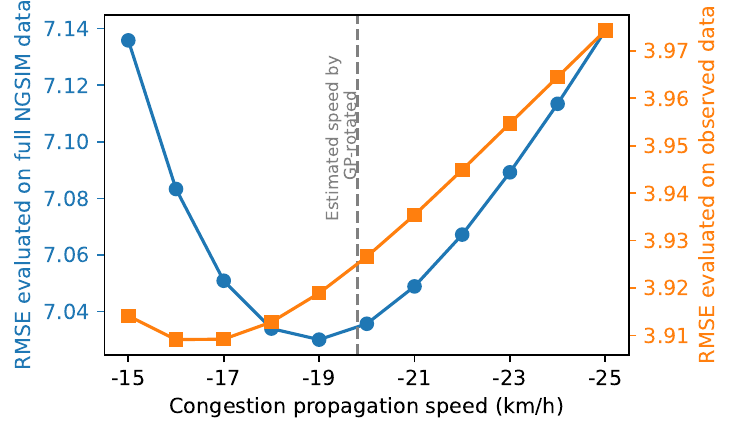}
    \caption{When observing 5\% trajectories, the RMSE error of ASM when using different congestion propagation speed in the NGSIM dataset.}
    \label{fig:rmse_speed}
\end{figure}

The GP-rotated kernel hyperparameters, $\alpha$, represent the angle between the traffic wave and the spatial direction, and values learned from the data are around $\alpha=0.108=6.20^{\circ}$ for the NGSIM dataset and $\alpha=0.160=9.16^{\circ}$ for the HighD dataset. After unit conversion with the size of cells, our estimation shows that the congestion propagation speed is approximately -19.87 km/h and -17.86 km/h for the NGSIM and HighD datasets, respectively, which is faster than the -15 km/h value used in \citet{treiber2011reconstructing}. { To ensure a fair comparison between ASM and GP-rotated, we set the congestion propagation speed of ASM in this study using the value derived by $\alpha$. Additionally, we assess the RMSE of TSE using ASM with various congestion propagation speeds, as depicted in Fig.\ref{fig:rmse_speed}. We observe that the speed estimated by GP-rotated (-19.87 km/h) is close to the optimal value (around -19 km/h) with minimum RMSE evaluated on the full NGSIM data (illustrated by the blue curve with dot marks). It's important to note that during the training phase, only 5\% of trajectories are observed. Consequently, while the best congestion propagation speed of ASM on the training set hovers around -17 km/h, it may not yield optimal performance for the full data. The insights from Fig.\ref{fig:rmse_speed} indicate that GP-rotated offers a promising approach for estimating congestion propagation speed from sparse vehicle trajectories.}

% \begin{itemize}
%     \item Our method performs the best under small observation ratio. But when the obersvation ratio is high STH-LRTC has the best performance. This is because the GP cannot capture the fine-grained texture in traffic flow, as shown in the Section~\ref{sec:uncertainty}.
%     \item When missing rate is high, the STH-LRTC could fail due to the large block of missing. This is when STH-LRTC has large mean error and STD...
%     \item GP-rotated is better than GP-ARD, this is because...
%     \item With the increase of ratio, the gap between GP-ARD and GP rotated narrows down.
%     \item We should further explain the meaning of the GP parameters, the length scale, theta.
% \end{itemize}

\subsection{TSE from loop detectors} \label{sec: loop performance}
The proposed TSE can also be used when traffic flow information is obtained from loop detectors. Loop detectors are commonly deployed at select road segments to gather specific traffic data, including vehicle count and density. Subsequently, these data points are leveraged to conduct traffic analysis. The installation of loop detectors is often sparse because of the high cost. Therefore, it is critical to utilize these sparse observations from loop detectors to perform TSE to reconstruct the complete traffic conditions on the whole road.

\begin{figure*}[!htb]
  \begin{center}
  \includegraphics[width=0.8\textwidth]{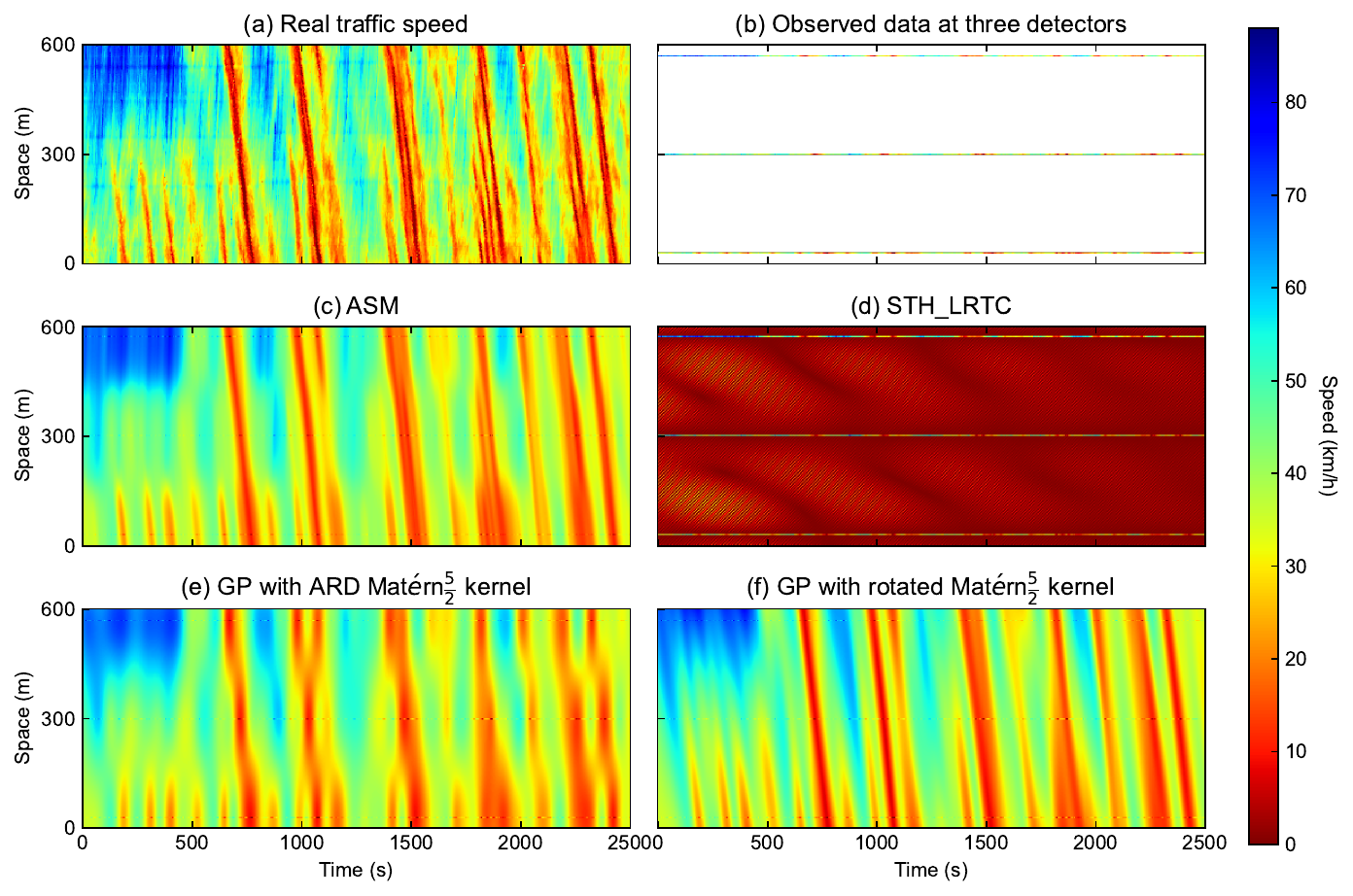}
  \caption{An TSE experiment on the NGSIM dataset with observation at three detectors.}
  \label{fig: traffic speed with detector}
  \end{center}
\end{figure*}

Our experiment commences by conducting TSE exclusively with loop detector data. For the NGSIM dataset, three virtual detectors are positioned at distances of 30 meters, 300 meters, and 570 meters from the starting point, as is shown in Fig.~\ref{fig: traffic speed with detector} (b). For the HighD dataset, the locations of the three detectors are 40 meters, 200 meters, and 360 meters from the starting point. Only the traffic speed at the location of the loop detectors can be observed.

Fig.~\ref{fig: traffic speed with detector} showcases the TSE results of the NGSIM dataset. Fig.~\ref{fig: traffic speed with detector} (a) presents the ground truth traffic speed map for the entire road segment. Fig.~\ref{fig: traffic speed with detector} (b) portrays the observed traffic states at the three specific locations where loop detectors are installed. Notably, STH-LRTC fails in this case because of blocks of missing rows in the matrix. Upon closer examination of Fig.~\ref{fig: traffic speed with detector} (c) and (f). {Both ASM and GP-rotated produce an accurate reconstruction of traffic speed, the success of these methods is largely attributed to the accurate estimation of the congestion propagation speed.} By comparing Fig.~\ref{fig: traffic speed with detector} (e) and (f), it is strikingly evident that our proposed GP-rotated method accurately captures congestion propagation and directional traffic wave correlations. This starkly contrasts with the traditional GP-ARD model, which fails to achieve the same level of precision in these aspects.

\begin{table}[!htbp]
  \centering
  \caption{TSE accuracy with observations at three detectors.}
   \setlength{\tabcolsep}{3.1mm}{
     \begin{tabular}{lcccc}
    \toprule
    Dataset & \multicolumn{2}{c}{NGSIM} & \multicolumn{2}{c}{HighD} \\
    \midrule
    Method & MAE   & RMSE  & MAE   & RMSE \\
    \midrule
    ASM   & { 4.45}  & {5.78}  & {3.05}  & {4.10}  \\
    STH-LRTC & 34.90  & 38.22  & 29.15  & 34.14  \\
    GP-ARD & 5.16  & 6.80  & 3.05  & 4.11  \\
    GP-rotated & \textbf{4.36}  & \textbf{5.59}  & \textbf{2.96}  & \textbf{4.01}  \\
    \bottomrule
    \end{tabular}}%
  \label{tab: performance with detectors}%
\end{table}%

Table~\ref{tab: performance with detectors} presents the MAE and RMSE values measured in m/s for each method applied to the two study areas. These findings align with the performance trends depicted in Fig.~\ref{fig: traffic speed with detector}. The proposed GP-rotated method outperforms the others when dealing with limited data obtained from specific loop detectors. For instance, when comparing the STH-LRTC method with GP-rotated, the former yields an MAE of 34.90 m/s and an RMSE of 38.22 m/s on the NGSIM dataset, whereas our proposed method achieves a significantly lower MSE of 4.36 m/s. This marked difference in performance is attributed to the incapacity of STH-LRTC to function effectively under such conditions. In comparing the GP-ARD and GP-rotated models, the latter demonstrates superior accuracy over the traditional GP model, with gaps in MAE (m/s) and RMSE (m/s) ranging from 0.1 to 1.2 m/s. Although the absolute gaps seem not vast, the relative difference cannot be ignored. It is vividly shown in Fig.~\ref{fig: traffic speed with detector}. { The ASM's accuracy surpasses that of GP-ARD but slightly lags behind GP-rotated. While it is possible to further enhance ASM performance by fine-tuning other parameters, such as the smoothing widths and the free-congested transition speed, this process is heavily dependent on prior knowledge; given that detectors offer only limited information to verify these parameters, achieving significant improvements for ASM is challenging. }

\subsection{TSE at a bottleneck}\label{sect: bottleneck}
\begin{figure*}[!ht]

  \begin{center}
  \includegraphics[width=0.8\textwidth]{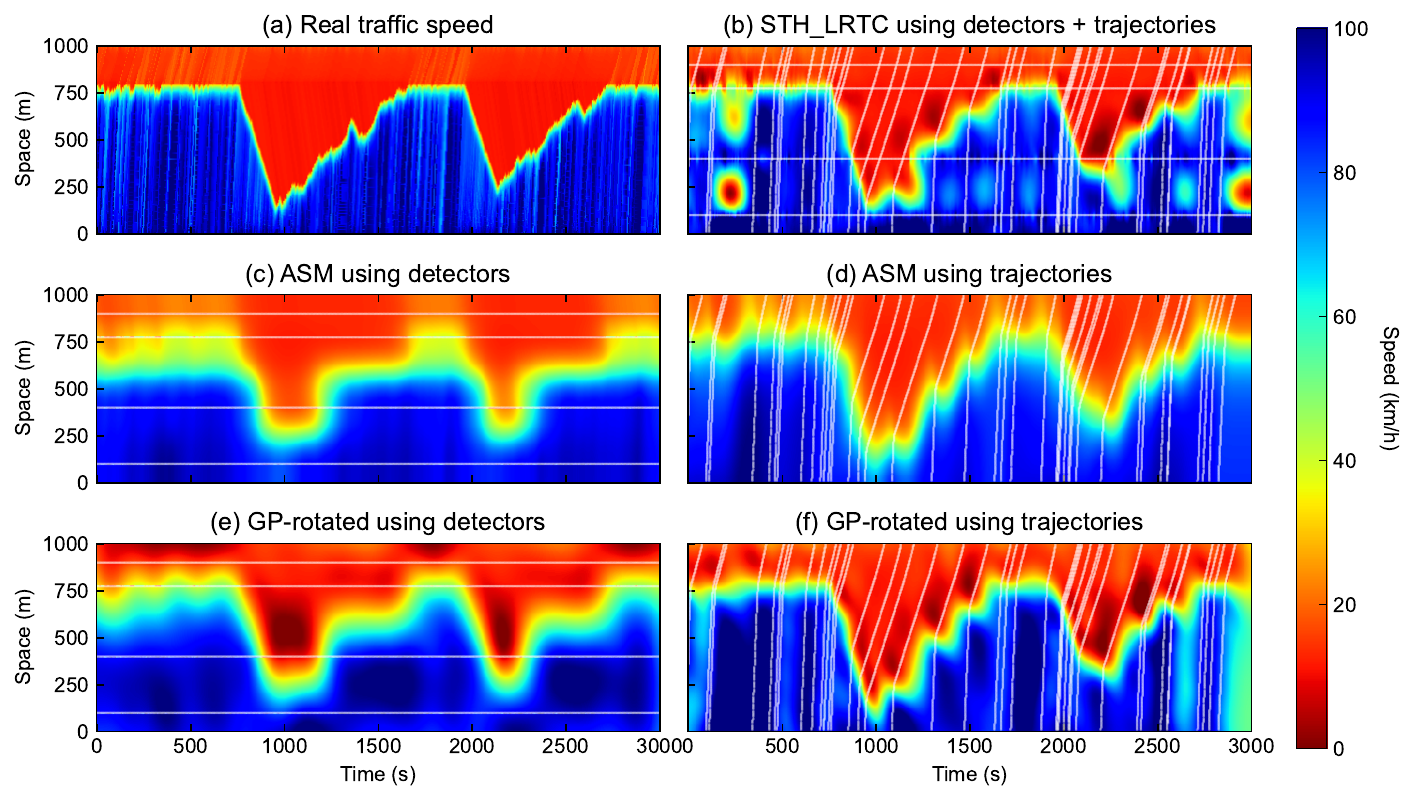}
  \caption{A TSE experiment on a simulated bottleneck. Locations of trajectories and detectors are shown in white lines. (a) The traffic speed of full simulated data. (b) The traffic speed of observed trajectories. (c) The traffic speed estimated by the ASM method. (d) The traffic speed estimated by the STH-LRTC method. (e) The traffic speed estimated by the GP with ARD Mat\'ern$\frac{5}{2}$ kernel. (f) The traffic speed estimated by the GP with the proposed rotated Mat\'ern$\frac{5}{2}$ kernel.}
  \label{fig:sim_traffic_speed}
  \end{center}
\end{figure*}

{ We further validate the performance of our proposed method using simulated traffic data in a bottleneck scenario to ensure its effectiveness in broader contexts. The simulation was executed in SUMO \citep[Simulation of Urban MObility,][]{lopez2018microscopic} with the Intelligent Driver Model \citep[IDM,][]{treiber2000congested} chosen as the car-following model. The total road length spans 1 km, with a speed limit of 100 km/h for the initial 750 m, followed by a bottleneck segment of 250 m with a speed limit reduced to 20 km/h. Vehicle arrivals follow a Poisson process with an expected rate of 720 veh/h. Additionally, we introduced two peak periods with vehicle demand set at 2160 veh/h and 1800 veh/h during the time intervals 800--1000 seconds and 2000--2200 seconds, respectively, to induce shockwaves. The simulation duration spans 3600 seconds, and we only focus on the midsection excluding the warm-up and end times. Average speeds are calculated using a 5-second $\times$ 5-meter grid, and the resulting speed profile is illustrated in Fig.~\ref{fig:sim_traffic_speed}~(a). We test the TSE performance using detector data, CV trajectories, or both data sources. The four detectors are located at 100 m, 400 m, 770 m, and 900 m, respectively. We randomly select the trajectories of 5\% vehicles for TSE. The location of detectors and trajectories are shown by white dots/lines in Fig.~\ref{fig:sim_traffic_speed}. We set $\tau_s=\tau_t=50$ for STH-LRTC. For ASM, the propagation speed of congestion is set as -10 km /h by measuring from the graph; the space smoothing width and time smoothing width are set as 100 m and 50 s, respectively.}

{
By comparing the estimation results in Fig.~\ref{fig:sim_traffic_speed} with Fig.~\ref{fig: traffic speed}, we observe that reconstructing the traffic state at a bottleneck presents greater challenges. This is attributed to congestion not only propagates backward but also forward as it returns to a free-flow state. Similar findings are discussed in Section 3.3 of the work by \citet{treiber2011reconstructing}; the detector data alone is not sufficient for accurately estimating the margins of shockwaves. Using more observations and a fusion of different data sources are required to obtain a good TSE for this complex scenario. Additionally, we note that the estimated rotation angle in GP-rotated no longer aligns with the backward propagation speed of congestion in this bottleneck example (the estimated angle corresponds to a speed ranging from -15 km/h to - 25 km/h, while the real value is around -10 km/h). This discrepancy arises from congestion propagation not being the predominant pattern in this scenario. Despite lacking a clear physical interpretation, GP-rotated achieves the best TSE performance in this experiment, as shown in Table~\ref{tab:sim_tse}, showing the versatility of GP-rotated for TSE across diverse scenarios.
}

% Table generated by Excel2LaTeX from sheet 'Sheet5'
\begin{table}[!htbp]

  \centering
  \caption{TSE accuracy for a simulated bottleneck with different data sources}
    \begin{tabular}{ccccccc}
    \toprule
    Data source & \multicolumn{2}{c}{Detectors} & \multicolumn{2}{c}{Trajectories} & \multicolumn{2}{c}{Detectors + trajectories} \\
    \midrule
    Method & RMSE  & MAE   & RMSE  & MAE   & RMSE  & MAE \\
    \midrule
    ASM   & 16.27 & 11.04 & 13.09 & 9.40  & 11.95 & 8.27 \\
    STH-LRTC & 54.97 & 42.05 & 22.27 & 12.51 & 15.07 & 8.25 \\
    GP-ARD & 19.34 & 13.83 & 16.80 & 11.46 & 12.51 & 8.99 \\
    GP-rotated & \textbf{13.40} & \textbf{9.55} & \textbf{12.67} & \textbf{8.45} & \textbf{9.75} & \textbf{6.27} \\
    \bottomrule
    \end{tabular}%
  \label{tab:sim_tse}%
\end{table}%

\subsection{Uncertainty quantification} \label{sec: uncertainty}

\begin{figure}
\begin{center}
\includegraphics[width=0.6\textwidth]{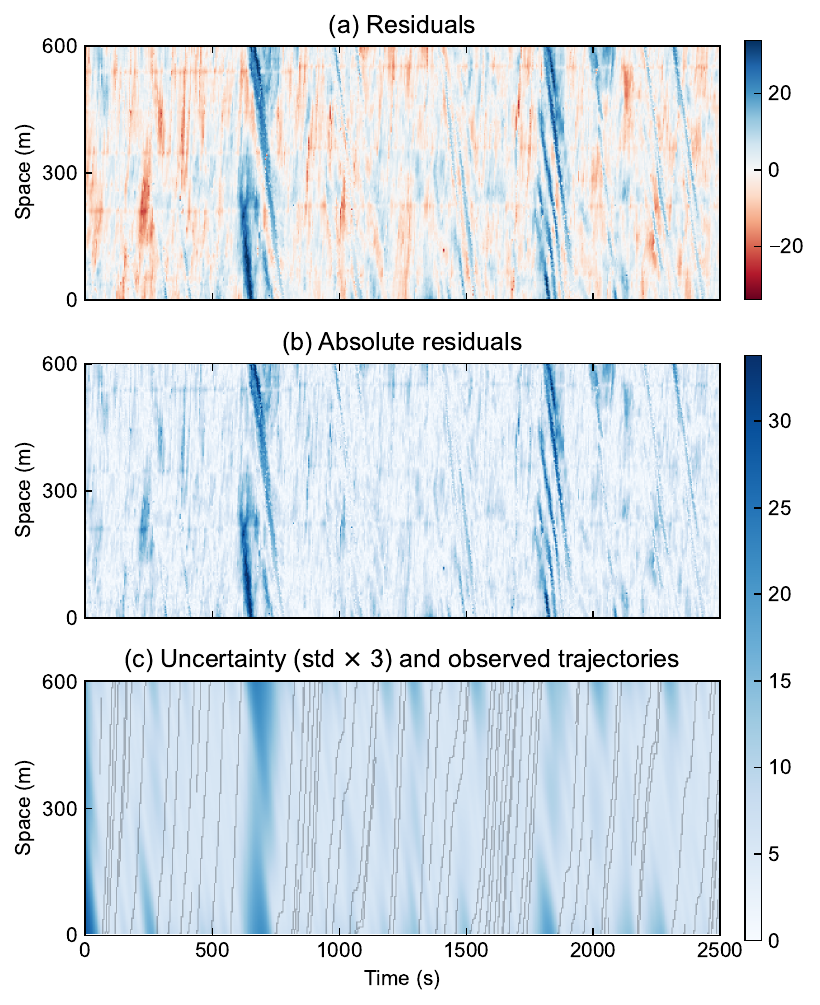}
\caption{The uncertainty quantification of the GP-rotated estimation method on the NGSIM dataset with 5\% observed trajectories. (a) The residuals of the estimation. (b) The absolute residuals of the estimation. (c) The uncertainty and 5\% observed trajectories.}
\label{fig: error}
\end{center}
\end{figure}

Our research incorporates uncertainty quantification as a crucial aspect to enable reliable and accurate predictions while acknowledging the inherent variability and unpredictability of the system under investigation, which is a lacking feature in existing methods. The GP framework provides a natural way to quantify uncertainty through the predictive covariance matrix, as shown in Eq.~\eqref{eq:cov_svgp}. We can use the diagonal elements of the covariance matrix (i.e., variance) to quantify the uncertainty of the TSE at each cell. The comparison between the TSE residuals and the uncertainty is presented in Fig.~\ref{fig: error}, which provides a comprehensive understanding of the uncertainties associated with our findings.

The uncertainties (shown by three standard deviations) of the TSE using GP-rotated and the observed trajectories are demonstrated in Fig.~\ref{fig: error} (c). First, we can find that the uncertainties are larger for regions with no CV trajectories, such as time ranges of 600 s to 700 s and 1800 s to 1900 s. It is notable that the uncertainty is anisotropic, propagating along the traffic wave, highlighting the need for an approach that can accurately capture this behavior. By comparing Fig.~\ref{fig: error} (b) and (c), we can find that the predictive uncertainties are, in general, consistent with the absolute residuals, meaning the predictive variance of GP-rotated is a reliable indicator for uncertainty quantification.

Finally, if we look at Fig.~\ref{fig: error} (a), we can find that there are still spatiotemporal correlations in the residuals, meaning that there is still space for improvement in the TSE estimation. For example, we can use the addition of multiple GP kernels, one for the congestion propagation and the other for the free-flow traffic, to capture the complex traffic dynamics. We have actually tried to use multiple GP kernels in our research, but the results do not improve. We believe that the reason is that the propagation of free flow speed is not as apparent as the congestion propagation, which does not provide enough information to improve the TSE estimation. But the correlations in the residuals still indicate that a more capable kernel design is needed to capture the complex traffic dynamics.

% traffic dynamics are not only affected by the congestion propagation but also by the traffic demand, which is not captured by the GP-rotated method. Therefore, we believe that the multiple GP kernels are not suitable for the TSE estimation.
% consistent with the trend in absolute residuals. The absolute residuals, which indicate the magnitude of the errors in our model predictions, are larger within the time ranges of 600 s to 700 s and 1800 s to 1900 s, corresponding to areas without trajectories shown in Fig.~\ref{fig: error} (c).
%  Our research shows that a single GP is not sufficient to capture traffic waves in TSE, as demonstrated in Fig.~\ref{fig: error} (a) and (b).
% Furthermore, the problem cannot be solved using the stationary GP method alone. Instead, our proposed GP-rotated method, which incorporates new kernel designs and mixing kernels, is required to address this issue.

In conclusion, our research emphasizes the importance of using an approach that can accurately capture the behavior of traffic waves in TSE. The GP-rotated method we propose is crucial in accounting for uncertainty propagation and allows us to provide reliable and accurate predictions. Through our approach, we can evaluate the validity of our model while also providing a measure of confidence in our predictions.

% \begin{itemize}
%     \item In general, the std of the GP-rotated prediction reflects trend in absolute residuals
%     \item The uncertainty is larger for areas without trajectories.
%     \item The uncertainty propagates along the traffic wave direction.
%     \item They are still correlations in the figure, (a), meaning that a single GP is still not sufficient. It could be solved by kernel design, mixing kernel, not stationary GP, etc.
% \end{itemize}

\subsection{Computational time} \label{sec: computation}

In Table~\ref{tab: computation}, we present the running time taken by four different methods, namely ASM, STH-LRTC, GP-rotated, and Pre-trained GP-rotated (P-GP-rotated), on both NGSIM and HighD datasets. Among these methods, the P-GP-rotated approach stands out for its significantly shorter computational time. This is because the P-GP-rotated method uses fixed kernel hyperparameters and random inducing points without any learning process. It is worth noting that ASM and STH-LRTC also use fixed parameters without an parameter estimation stage, which makes it appropriate to compare them with P-GP-rotated rather than GP-rotated.

We can see the running time in the highD dataset is faster than the NGSIM dataset. This is because the highD dataset has a smaller grid size. The computational time of STH-LRTC is considerably higher compared to other methods. For instance, on HighD data, it takes approximately 20 to 60 times longer than the P-GP-rotated method and 15 to 30 times longer than ASM computation. Moreover, the computational efficiency of STH-LRTC drops significantly with a lower the penetration rate. This is mainly due to the increase in the spatiotemporal delay embedding lengths ($\tau_s$ and $\tau_t$), which impacts the computation time substantially. As a result, the computational cost of STH-LRTC becomes extremely high under such scenarios. However, it is essential to note that this trend might not always hold, and a slight change in the parameters of the delay embedding in STH-LRTC could alter the trend.

We observe that the computational time of ASM, GP-rotated, and P-GP-rotated methods increases as the penetration rate increases. This is understandable as more data needs to be processed, leading to higher computation costs due to the increased traffic information. { It's worth highlighting that both ASM and GP-based methods can benefit from using a locality approximation that excludes distant points in the filters/covariance matrices to speed up the computation \citep[e.g.,][]{gramacy2015local}. Besides, our testing only employs a naive implementation of ASM. Faster implementations of ASM exist, leveraging efficient matrix operations and the Fast Fourier Transform (FFT) \citep{schreiter2010two}, which can reduce computation time by two orders of magnitude. Considering these, while P-GP-rotated demonstrates satisfactory computational efficiency, ASM can be significantly faster with proper implementations.
}

\begin{table}[htbp]
  \centering
  \caption{Computational time in seconds, mean (std).}
  \begin{threeparttable}
    \begin{tabular}{ccccc}
    \toprule
    \multicolumn{5}{c}{NGSIM} \\
    \midrule
    Rate  & ASM   & STH-LRTC & GP-rotated & P-GP-rotated \\
    \midrule
    0.05  & 7.40 (0.57) & 908.21 (38.61)$^{a}$ & 27.30 (2.92) & \textbf{3.84 (0.27)} \\
    0.1   & 14.18 (0.43) & 850.90 (19.61)$^{a}$ & 77.54 (4.68) & \textbf{9.25 (0.42)} \\
    0.2   & 26.77 (0.92) & 206.72 (1.85) & 153.07 (3.67) & \textbf{13.43 (1.68)} \\
    0.3   & 38.38 (1.38) & 199.99 (1.77) & 204.61 (2.71) & \textbf{13.97 (0.19)} \\
    0.4   & 48.15 (3.98) & 196.09 (2.66) & 245.37 (3.76) & \textbf{14.94 (0.16)} \\
    0.5   & 54.21 (2.83) & 191.46 (1.93) & 280.01 (5.28) & \textbf{15.85 (0.26)} \\
    \midrule
    \multicolumn{5}{c}{HighD} \\
    \midrule
    Rate  & ASM   & STH-LRTC & GP-rotated & P-GP-rotated \\
    \midrule
     0.05  & 0.46 (0.03) & 67.61 (3.38) & 11.97 (1.89) & \textbf{0.35 (0.05)} \\
    0.1   & 0.87 (0.04) & 823.65 (10.74)$^{b}$ & 12.54 (0.29) & \textbf{0.42 (0.05)} \\
    0.2   & 1.67 (0.07) & 54.29 (1.39) & 19.86 (0.40) & \textbf{0.83 (0.16)} \\
    0.3   & 2.30 (0.05) & 51.46 (1.06) & 29.78 (0.66) & \textbf{1.25 (0.15)} \\
    0.4   & 2.84 (0.08) & 49.50 (0.89) & 40.46 (0.86) & \textbf{1.63 (0.21)} \\
    0.5   & 3.22 (0.09) & 48.14 (0.82) & 50.93 (1.49) & \textbf{2.27 (0.21)} \\
    \bottomrule
    \end{tabular}%
    \begin{tablenotes}
      \item[a] Delay-embedding lengths $\tau_s=50$, $\tau_t=50$.
      \item[b] Delay-embedding lengths $\tau_s=60$, $\tau_t=50$.
  \end{tablenotes}
  \end{threeparttable}%
  \label{tab: computation}%
\end{table}%

\subsection{TSE in multiple lanes} \label{sec: case multi-lane}
\begin{figure*}[!ht]
	\begin{center}
	\includegraphics[width=0.8\textwidth]{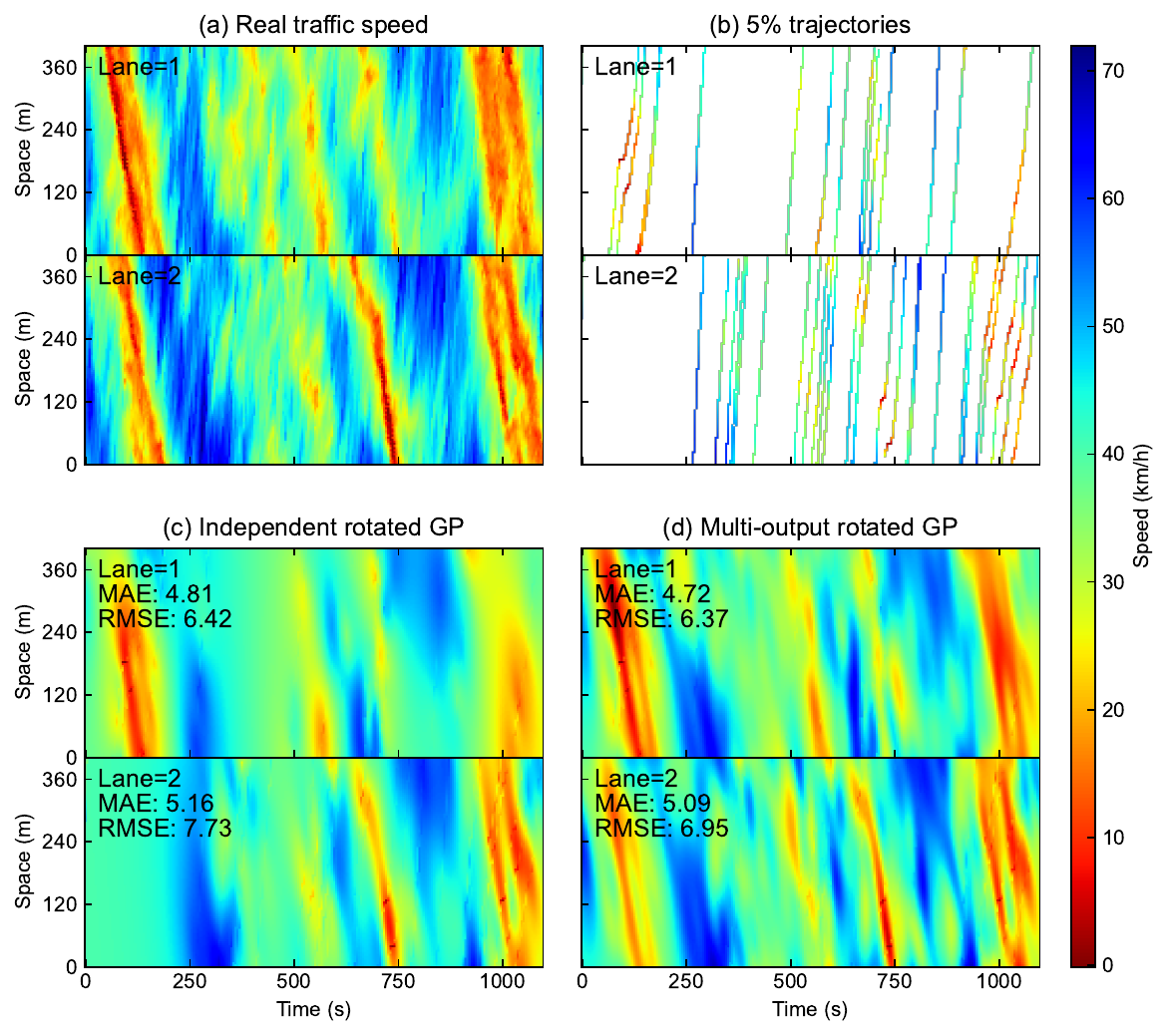}
	\caption{The TSE in multiple lanes on the HighD dataset. { TSE errors are marked on the top-left corner.}. (a) The traffic speed profile of the full dataset. (b) The traffic speed profile of the 5\% observed trajectories. (c) The traffic state is estimated by the independent rotated GP method. (d) The traffic state is estimated by the mul rotated GP method.}
	\label{fig:icm}
	\end{center}
\end{figure*}

Most previous works have focused on modeling the TSE of each lane independently without considering the correlations and interactions between neighboring lanes. { This section demonstrates the multi-output GP introduced in Section~\ref{sec: multi-output GP} that enhances the TSE by learning the correlations of traffic states on multiple lanes.} Specifically, we use the trajectories from all lanes as the input features to the multi-output GP model and predict the traffic state for each lane as a separate output dimension. By sharing the same covariance structure across all output dimensions, the multi-output GP model can capture the correlations and dependencies between the traffic states on different lanes, bringing more accurate and robust predictions.

The correlation between traffic speed profiles on different lanes of a highway segment is evident from Fig.~\ref{fig:icm} (a). The ground truth traffic speed profiles for Lane 1 and Lane 2 demonstrate that these two lanes are correlated, with both lanes experiencing congestion during the first 200 seconds and the last 100 seconds of the observed period. From 250 s to 350 s, the traffic speeds on both lanes are high. However, when we have only 5\% of trajectory data for each lane, accurately and simultaneously estimating the traffic state on these two lanes becomes challenging.

In Fig.~\ref{fig:icm} (b), it can be observed that either Lane 1 or Lane 2 has a gap without any observations, lasting from 260 s to 490 s on lane 1 and from 0 to 250 s on lane 2, respectively. When using the independent GP-rotated method to perform TSE, the resulting traffic state profiles are shown in Fig.~\ref{fig:icm} (c). It is evident from the figure that the speed map of Lane 1 does not contain much traffic information during the time period between 250 s and 500 s, while the speed map of Lane 2 loses most of the traffic information between 0 s and 250 s. These results demonstrate that performing independent TSE in each lane cannot achieve high estimation accuracy.

Despite the long periods of missing observations in either Lane 1 or Lane 2, we can observe that the trajectories from the other lane can compensate for the missing period, which is the main idea behind using a multi-output GP to output TSE in multiple lanes. Fig.~\ref{fig:icm} (d) shows that the multi-output GP performs better than the independent GP method. Specifically, during the period between 0 s and 250 s, the multi-output GP can reconstruct the shockwave of Lane 2 using the information from Lane 1, whereas the independent rotated GP fails to rebuild this shockwave. { As shown in the top-left corner of Fig.~\ref{fig:icm} (c) and (d), the MAE and RMSE error of multi-output rotated GP is significantly smaller than independent rotated GP.}

\section{Conclusion and Discussion}
\label{sec: conclusion}
This paper presents a novel approach for traffic speed estimation using Gaussian process regression with a rotated
kernel parametrization. The rotated kernel is designed to model anisotropic traffic flow, allowing for capturing the directional dependence of traffic wave propagation. The proposed method is a generalization of the ARD kernel function and can be applied to other kernel functions like Mat\'ern and rational quadratic kernels. To validate the effectiveness of the proposed method, we conduct experiments on two real-world datasets from the NGSIM and HighD programs, {as well as a simulated dataset for a traffic bottleneck scenario}. { The findings from comprehensive experiments underscore the capability of the GP-rotated TSE method across varying CV penetration rates and detector types. Notably, our approach achieves state-of-the-art accuracy in scenarios with sparse CV penetration rates. Moreover, GP-rotated offers a promising approach for estimating congestion propagation speed from sparse vehicle trajectories.} The outputs of GP-rotated also provide statistical uncertainty quantification, which is crucial for data-driven TSE models, especially under limited training data. { We also extend GP-rotated to capture the speed correlations of multiple lanes, significantly improving the TSE accuracy for multiple lanes with sparse observations.} Overall, the proposed method is a promising approach for traffic speed estimation, offering improved performance and the ability to capture directional traffic flow patterns.

While the proposed method shows promising results in traffic speed estimation, there are some limitations and potential future research directions. First, the current model is only tested on the traffic speed estimation problem, and it may be possible to estimate speed, density, and other traffic state variables simultaneously using multi-output Gaussian process regression. Second, future research can extend the model to assess the traffic wave by incorporating additional information, such as traffic signals and road geometry, to make the model suitable for more scenarios. Third, the proposed method is evaluated on real-world trajectory datasets that simulate the trajectory data obtained from connected vehicles. To further validate the effectiveness of the proposed method, it is suggested to test on commercial CV datasets.

\section*{Acknowledgment}
The authors would like to thank the Natural Sciences and Engineering Research Council (NSERC) of Canada Industrial Research Chair (IRC) Grant for funding support. The contents of this paper reflect the views of the authors, who are responsible for the facts and the accuracy of the data presented herein. This paper does not constitute a standard, specification, or regulation. The code and data associated with this paper are available at \url{https://github.com/Lucky-Fan/GP_TSE}.

\bibliographystyle{cas-model2-names}

\bibliography{ref}

\begin{thebibliography}{60}
\expandafter\ifx\csname natexlab\endcsname\relax\def\natexlab#1{#1}\fi
\providecommand{\url}[1]{\texttt{#1}}
\providecommand{\href}[2]{#2}
\providecommand{\path}[1]{#1}
\providecommand{\DOIprefix}{doi:}
\providecommand{\ArXivprefix}{arXiv:}
\providecommand{\URLprefix}{URL: }
\providecommand{\Pubmedprefix}{pmid:}
\providecommand{\doi}[1]{\href{http://dx.doi.org/#1}{\path{#1}}}
\providecommand{\Pubmed}[1]{\href{pmid:#1}{\path{#1}}}
\providecommand{\bibinfo}[2]{#2}
\ifx\xfnm\relax \def\xfnm[#1]{\unskip,\space#1}\fi
%Type = Article
\bibitem[{Aw and Rascle(2000)}]{aw2000resurrection}
\bibinfo{author}{Aw, A.}, \bibinfo{author}{Rascle, M.}, \bibinfo{year}{2000}.
\newblock \bibinfo{title}{Resurrection of" second order" models of traffic
  flow}.
\newblock \bibinfo{journal}{SIAM journal on applied mathematics}
  \bibinfo{volume}{60}, \bibinfo{pages}{916--938}.
%Type = Article
\bibitem[{Bekiaris-Liberis et~al.(2017)Bekiaris-Liberis, Roncoli and
  Papageorgiou}]{bekiaris2017highway}
\bibinfo{author}{Bekiaris-Liberis, N.}, \bibinfo{author}{Roncoli, C.},
  \bibinfo{author}{Papageorgiou, M.}, \bibinfo{year}{2017}.
\newblock \bibinfo{title}{Highway traffic state estimation per lane in the
  presence of connected vehicles}.
\newblock \bibinfo{journal}{Transportation research part B: methodological}
  \bibinfo{volume}{106}, \bibinfo{pages}{1--28}.
%Type = Article
\bibitem[{Bonilla et~al.(2007)Bonilla, Chai and Williams}]{bonilla2007multi}
\bibinfo{author}{Bonilla, E.V.}, \bibinfo{author}{Chai, K.},
  \bibinfo{author}{Williams, C.}, \bibinfo{year}{2007}.
\newblock \bibinfo{title}{Multi-task gaussian process prediction}.
\newblock \bibinfo{journal}{Advances in neural information processing systems}
  \bibinfo{volume}{20}.
%Type = Inproceedings
\bibitem[{Chen and Levin(2019)}]{chen2019traffic}
\bibinfo{author}{Chen, R.}, \bibinfo{author}{Levin, M.W.},
  \bibinfo{year}{2019}.
\newblock \bibinfo{title}{Traffic state estimation based on kalman filter
  technique using connected vehicle v2v basic safety messages}, in:
  \bibinfo{booktitle}{2019 IEEE Intelligent Transportation Systems Conference
  (ITSC)}, \bibinfo{organization}{IEEE}. pp. \bibinfo{pages}{4380--4385}.
%Type = Article
\bibitem[{Cheng et~al.(2022)Cheng, Wang, Chen, Tr{\'e}panier and
  Sun}]{cheng2022bayesian}
\bibinfo{author}{Cheng, Z.}, \bibinfo{author}{Wang, X.}, \bibinfo{author}{Chen,
  X.}, \bibinfo{author}{Tr{\'e}panier, M.}, \bibinfo{author}{Sun, L.},
  \bibinfo{year}{2022}.
\newblock \bibinfo{title}{Bayesian calibration of traffic flow fundamental
  diagrams using gaussian processes}.
\newblock \bibinfo{journal}{IEEE Open Journal of Intelligent Transportation
  Systems} \bibinfo{volume}{3}, \bibinfo{pages}{763--771}.
%Type = Article
\bibitem[{Florin and Olariu(2016)}]{florin2016variant}
\bibinfo{author}{Florin, R.}, \bibinfo{author}{Olariu, S.},
  \bibinfo{year}{2016}.
\newblock \bibinfo{title}{On a variant of the mobile observer method}.
\newblock \bibinfo{journal}{IEEE Transactions on Intelligent Transportation
  Systems} \bibinfo{volume}{18}, \bibinfo{pages}{441--449}.
%Type = Article
\bibitem[{Fountoulakis et~al.(2017)Fountoulakis, Bekiaris-Liberis, Roncoli,
  Papamichail and Papageorgiou}]{fountoulakis2017highway}
\bibinfo{author}{Fountoulakis, M.}, \bibinfo{author}{Bekiaris-Liberis, N.},
  \bibinfo{author}{Roncoli, C.}, \bibinfo{author}{Papamichail, I.},
  \bibinfo{author}{Papageorgiou, M.}, \bibinfo{year}{2017}.
\newblock \bibinfo{title}{Highway traffic state estimation with mixed connected
  and conventional vehicles: Microscopic simulation-based testing}.
\newblock \bibinfo{journal}{Transportation Research Part C: Emerging
  Technologies} \bibinfo{volume}{78}, \bibinfo{pages}{13--33}.
%Type = Article
\bibitem[{Gramacy and Apley(2015)}]{gramacy2015local}
\bibinfo{author}{Gramacy, R.B.}, \bibinfo{author}{Apley, D.W.},
  \bibinfo{year}{2015}.
\newblock \bibinfo{title}{Local gaussian process approximation for large
  computer experiments}.
\newblock \bibinfo{journal}{Journal of Computational and Graphical Statistics}
  \bibinfo{volume}{24}, \bibinfo{pages}{561--578}.
%Type = Article
\bibitem[{Han and Ahn(2021)}]{han2021estimation}
\bibinfo{author}{Han, Y.}, \bibinfo{author}{Ahn, S.}, \bibinfo{year}{2021}.
\newblock \bibinfo{title}{Estimation of traffic flow rate with data from
  connected-automated vehicles using bayesian inference and deep learning}.
\newblock \bibinfo{journal}{Frontiers in Future Transportation}
  \bibinfo{volume}{2}, \bibinfo{pages}{644988}.
%Type = Inproceedings
\bibitem[{Jia et~al.(2016)Jia, Wu and Du}]{jia2016traffic}
\bibinfo{author}{Jia, Y.}, \bibinfo{author}{Wu, J.}, \bibinfo{author}{Du, Y.},
  \bibinfo{year}{2016}.
\newblock \bibinfo{title}{Traffic speed prediction using deep learning method},
  in: \bibinfo{booktitle}{2016 IEEE 19th international conference on
  intelligent transportation systems (ITSC)}, \bibinfo{organization}{IEEE}. pp.
  \bibinfo{pages}{1217--1222}.
%Type = Article
\bibitem[{Kerner(1999)}]{kerner1999physics}
\bibinfo{author}{Kerner, B.S.}, \bibinfo{year}{1999}.
\newblock \bibinfo{title}{The physics of traffic}.
\newblock \bibinfo{journal}{Physics World} \bibinfo{volume}{12},
  \bibinfo{pages}{25}.
%Type = Inproceedings
\bibitem[{Krajewski et~al.(2018)Krajewski, Bock, Kloeker and
  Eckstein}]{highDdataset}
\bibinfo{author}{Krajewski, R.}, \bibinfo{author}{Bock, J.},
  \bibinfo{author}{Kloeker, L.}, \bibinfo{author}{Eckstein, L.},
  \bibinfo{year}{2018}.
\newblock \bibinfo{title}{The highd dataset: A drone dataset of naturalistic
  vehicle trajectories on german highways for validation of highly automated
  driving systems}, in: \bibinfo{booktitle}{2018 21st International Conference
  on Intelligent Transportation Systems (ITSC)}, pp.
  \bibinfo{pages}{2118--2125}.
\newblock \DOIprefix\doi{10.1109/ITSC.2018.8569552}.
%Type = Article
\bibitem[{Kyriacou et~al.(2022)Kyriacou, Englezou, Panayiotou and
  Timotheou}]{kyriacou2022bayesian}
\bibinfo{author}{Kyriacou, V.}, \bibinfo{author}{Englezou, Y.},
  \bibinfo{author}{Panayiotou, C.G.}, \bibinfo{author}{Timotheou, S.},
  \bibinfo{year}{2022}.
\newblock \bibinfo{title}{Bayesian traffic state estimation using extended
  floating car data}.
\newblock \bibinfo{journal}{IEEE Transactions on Intelligent Transportation
  Systems} .
%Type = Article
\bibitem[{Li et~al.(2013)Li, Li and Li}]{li2013efficient}
\bibinfo{author}{Li, L.}, \bibinfo{author}{Li, Y.}, \bibinfo{author}{Li, Z.},
  \bibinfo{year}{2013}.
\newblock \bibinfo{title}{Efficient missing data imputing for traffic flow by
  considering temporal and spatial dependence}.
\newblock \bibinfo{journal}{Transportation research part C: emerging
  technologies} \bibinfo{volume}{34}, \bibinfo{pages}{108--120}.
%Type = Article
\bibitem[{Lighthill and Whitham(1955)}]{lighthill1955kinematic}
\bibinfo{author}{Lighthill, M.J.}, \bibinfo{author}{Whitham, G.B.},
  \bibinfo{year}{1955}.
\newblock \bibinfo{title}{On kinematic waves ii. a theory of traffic flow on
  long crowded roads}.
\newblock \bibinfo{journal}{Proceedings of the Royal Society of London. Series
  A. Mathematical and Physical Sciences} \bibinfo{volume}{229},
  \bibinfo{pages}{317--345}.
%Type = Article
\bibitem[{Liu et~al.(2023)Liu, Lyu, Wang, Wang, Liu and Meng}]{liu2023gaussian}
\bibinfo{author}{Liu, Z.}, \bibinfo{author}{Lyu, C.}, \bibinfo{author}{Wang,
  Z.}, \bibinfo{author}{Wang, S.}, \bibinfo{author}{Liu, P.},
  \bibinfo{author}{Meng, Q.}, \bibinfo{year}{2023}.
\newblock \bibinfo{title}{A gaussian-process-based data-driven traffic flow
  model and its application in road capacity analysis}.
\newblock \bibinfo{journal}{IEEE Transactions on Intelligent Transportation
  Systems} \bibinfo{volume}{24}, \bibinfo{pages}{1544--1563}.
%Type = Inproceedings
\bibitem[{Lopez et~al.(2018)Lopez, Behrisch, Bieker-Walz, Erdmann,
  Fl{\"o}tter{\"o}d, Hilbrich, L{\"u}cken, Rummel, Wagner and
  Wie{\ss}ner}]{lopez2018microscopic}
\bibinfo{author}{Lopez, P.A.}, \bibinfo{author}{Behrisch, M.},
  \bibinfo{author}{Bieker-Walz, L.}, \bibinfo{author}{Erdmann, J.},
  \bibinfo{author}{Fl{\"o}tter{\"o}d, Y.P.}, \bibinfo{author}{Hilbrich, R.},
  \bibinfo{author}{L{\"u}cken, L.}, \bibinfo{author}{Rummel, J.},
  \bibinfo{author}{Wagner, P.}, \bibinfo{author}{Wie{\ss}ner, E.},
  \bibinfo{year}{2018}.
\newblock \bibinfo{title}{Microscopic traffic simulation using sumo}, in:
  \bibinfo{booktitle}{2018 21st international conference on intelligent
  transportation systems (ITSC)}, \bibinfo{organization}{IEEE}. pp.
  \bibinfo{pages}{2575--2582}.
%Type = Article
\bibitem[{Makridis and Kouvelas(2023)}]{makridis2023adaptive}
\bibinfo{author}{Makridis, M.A.}, \bibinfo{author}{Kouvelas, A.},
  \bibinfo{year}{2023}.
\newblock \bibinfo{title}{An adaptive framework for real-time freeway traffic
  estimation in the presence of cavs}.
\newblock \bibinfo{journal}{Transportation Research Part C: Emerging
  Technologies} \bibinfo{volume}{149}, \bibinfo{pages}{104066}.
%Type = Article
\bibitem[{Mihaylova et~al.(2006)Mihaylova, Boel and
  Hegyi}]{mihaylova2006unscented}
\bibinfo{author}{Mihaylova, L.}, \bibinfo{author}{Boel, R.},
  \bibinfo{author}{Hegyi, A.}, \bibinfo{year}{2006}.
\newblock \bibinfo{title}{An unscented kalman filter for freeway traffic
  estimation}.
\newblock \bibinfo{journal}{IFAC Proceedings Volumes} \bibinfo{volume}{39},
  \bibinfo{pages}{31--36}.
%Type = Article
\bibitem[{Mihaylova et~al.(2007)Mihaylova, Boel and
  Hegyi}]{mihaylova2007freeway}
\bibinfo{author}{Mihaylova, L.}, \bibinfo{author}{Boel, R.},
  \bibinfo{author}{Hegyi, A.}, \bibinfo{year}{2007}.
\newblock \bibinfo{title}{Freeway traffic estimation within particle filtering
  framework}.
\newblock \bibinfo{journal}{Automatica} \bibinfo{volume}{43},
  \bibinfo{pages}{290--300}.
%Type = Article
\bibitem[{Nanthawichit et~al.(2003)Nanthawichit, Nakatsuji and
  Suzuki}]{nanthawichit2003application}
\bibinfo{author}{Nanthawichit, C.}, \bibinfo{author}{Nakatsuji, T.},
  \bibinfo{author}{Suzuki, H.}, \bibinfo{year}{2003}.
\newblock \bibinfo{title}{Application of probe-vehicle data for real-time
  traffic-state estimation and short-term travel-time prediction on a freeway}.
\newblock \bibinfo{journal}{Transportation research record}
  \bibinfo{volume}{1855}, \bibinfo{pages}{49--59}.
%Type = Article
\bibitem[{NEAL(1996)}]{neal1996bayesian}
\bibinfo{author}{NEAL, R.}, \bibinfo{year}{1996}.
\newblock \bibinfo{title}{Bayesian learning for neural networks}.
\newblock \bibinfo{journal}{Lecture Notes in Statistics} .
%Type = Article
\bibitem[{NGSIM(2007)}]{simulation2007us}
\bibinfo{author}{NGSIM}, \bibinfo{year}{2007}.
\newblock \bibinfo{title}{Us highway 101 dataset} \URLprefix
  \url{https://www.fhwa.dot.gov/publications/research/operations/07030/index.cfm}.
%Type = Article
\bibitem[{Ni and Leonard(2005)}]{ni2005markov}
\bibinfo{author}{Ni, D.}, \bibinfo{author}{Leonard, J.D.},
  \bibinfo{year}{2005}.
\newblock \bibinfo{title}{Markov chain monte carlo multiple imputation using
  bayesian networks for incomplete intelligent transportation systems data}.
\newblock \bibinfo{journal}{Transportation research record}
  \bibinfo{volume}{1935}, \bibinfo{pages}{57--67}.
%Type = Article
\bibitem[{Payne(1971)}]{payne1971model}
\bibinfo{author}{Payne, H.J.}, \bibinfo{year}{1971}.
\newblock \bibinfo{title}{Model of freeway traffic and control}.
\newblock \bibinfo{journal}{Mathematical Model of Public System} ,
  \bibinfo{pages}{51--61}.
%Type = Book
\bibitem[{Rasmussen et~al.(2006)Rasmussen, Williams
  et~al.}]{rasmussen2006gaussian}
\bibinfo{author}{Rasmussen, C.E.}, \bibinfo{author}{Williams, C.K.}, et~al.,
  \bibinfo{year}{2006}.
\newblock \bibinfo{title}{Gaussian processes for machine learning}.
  volume~\bibinfo{volume}{1}.
\newblock \bibinfo{publisher}{Springer}.
%Type = Article
\bibitem[{Rempe et~al.(2022)Rempe, Franeck and
  Bogenberger}]{rempe2022estimation}
\bibinfo{author}{Rempe, F.}, \bibinfo{author}{Franeck, P.},
  \bibinfo{author}{Bogenberger, K.}, \bibinfo{year}{2022}.
\newblock \bibinfo{title}{On the estimation of traffic speeds with deep
  convolutional neural networks given probe data}.
\newblock \bibinfo{journal}{Transportation research part C: emerging
  technologies} \bibinfo{volume}{134}, \bibinfo{pages}{103448}.
%Type = Article
\bibitem[{Rempe et~al.(2017)Rempe, Franeck, Fastenrath and
  Bogenberger}]{rempe2017phase}
\bibinfo{author}{Rempe, F.}, \bibinfo{author}{Franeck, P.},
  \bibinfo{author}{Fastenrath, U.}, \bibinfo{author}{Bogenberger, K.},
  \bibinfo{year}{2017}.
\newblock \bibinfo{title}{A phase-based smoothing method for accurate traffic
  speed estimation with floating car data}.
\newblock \bibinfo{journal}{Transportation Research Part C: Emerging
  Technologies} \bibinfo{volume}{85}, \bibinfo{pages}{644--663}.
%Type = Article
\bibitem[{Richards(1956)}]{richards1956shock}
\bibinfo{author}{Richards, P.I.}, \bibinfo{year}{1956}.
\newblock \bibinfo{title}{Shock waves on the highway}.
\newblock \bibinfo{journal}{Operations research} \bibinfo{volume}{4},
  \bibinfo{pages}{42--51}.
%Type = Inproceedings
\bibitem[{Schreiter et~al.(2010)Schreiter, van Lint, Treiber and
  Hoogendoorn}]{schreiter2010two}
\bibinfo{author}{Schreiter, T.}, \bibinfo{author}{van Lint, H.},
  \bibinfo{author}{Treiber, M.}, \bibinfo{author}{Hoogendoorn, S.},
  \bibinfo{year}{2010}.
\newblock \bibinfo{title}{Two fast implementations of the adaptive smoothing
  method used in highway traffic state estimation}, in:
  \bibinfo{booktitle}{13th International IEEE Conference on Intelligent
  Transportation Systems}, \bibinfo{organization}{IEEE}. pp.
  \bibinfo{pages}{1202--1208}.
%Type = Article
\bibitem[{Seo et~al.(2017)Seo, Bayen, Kusakabe and Asakura}]{seo2017traffic}
\bibinfo{author}{Seo, T.}, \bibinfo{author}{Bayen, A.M.},
  \bibinfo{author}{Kusakabe, T.}, \bibinfo{author}{Asakura, Y.},
  \bibinfo{year}{2017}.
\newblock \bibinfo{title}{Traffic state estimation on highway: A comprehensive
  survey}.
\newblock \bibinfo{journal}{Annual reviews in control} \bibinfo{volume}{43},
  \bibinfo{pages}{128--151}.
%Type = Article
\bibitem[{Seo and Kusakabe(2015)}]{seo2015probe}
\bibinfo{author}{Seo, T.}, \bibinfo{author}{Kusakabe, T.},
  \bibinfo{year}{2015}.
\newblock \bibinfo{title}{Probe vehicle-based traffic state estimation method
  with spacing information and conservation law}.
\newblock \bibinfo{journal}{Transportation Research Part C: Emerging
  Technologies} \bibinfo{volume}{59}, \bibinfo{pages}{391--403}.
%Type = Article
\bibitem[{Seo et~al.(2015a)Seo, Kusakabe and Asakura}]{seo2015estimation}
\bibinfo{author}{Seo, T.}, \bibinfo{author}{Kusakabe, T.},
  \bibinfo{author}{Asakura, Y.}, \bibinfo{year}{2015}a.
\newblock \bibinfo{title}{Estimation of flow and density using probe vehicles
  with spacing measurement equipment}.
\newblock \bibinfo{journal}{Transportation Research Part C: Emerging
  Technologies} \bibinfo{volume}{53}, \bibinfo{pages}{134--150}.
%Type = Inproceedings
\bibitem[{Seo et~al.(2015b)Seo, Kusakabe and Asakura}]{seo2015traffic}
\bibinfo{author}{Seo, T.}, \bibinfo{author}{Kusakabe, T.},
  \bibinfo{author}{Asakura, Y.}, \bibinfo{year}{2015}b.
\newblock \bibinfo{title}{Traffic state estimation with the advanced probe
  vehicles using data assimilation}, in: \bibinfo{booktitle}{2015 IEEE 18th
  International Conference on Intelligent Transportation Systems},
  \bibinfo{organization}{IEEE}. pp. \bibinfo{pages}{824--830}.
%Type = Inproceedings
\bibitem[{Shi et~al.(2021a)Shi, Mo and Di}]{shi2021physics2}
\bibinfo{author}{Shi, R.}, \bibinfo{author}{Mo, Z.}, \bibinfo{author}{Di, X.},
  \bibinfo{year}{2021}a.
\newblock \bibinfo{title}{Physics-informed deep learning for traffic state
  estimation: A hybrid paradigm informed by second-order traffic models}, in:
  \bibinfo{booktitle}{Proceedings of the AAAI Conference on Artificial
  Intelligence}, pp. \bibinfo{pages}{540--547}.
%Type = Article
\bibitem[{Shi et~al.(2021b)Shi, Mo, Huang, Di and Du}]{shi2021physics}
\bibinfo{author}{Shi, R.}, \bibinfo{author}{Mo, Z.}, \bibinfo{author}{Huang,
  K.}, \bibinfo{author}{Di, X.}, \bibinfo{author}{Du, Q.},
  \bibinfo{year}{2021}b.
\newblock \bibinfo{title}{A physics-informed deep learning paradigm for traffic
  state and fundamental diagram estimation}.
\newblock \bibinfo{journal}{IEEE Transactions on Intelligent Transportation
  Systems} \bibinfo{volume}{23}, \bibinfo{pages}{11688--11698}.
%Type = Article
\bibitem[{Storm et~al.(2022)Storm, Mandjes and van Arem}]{storm2022efficient}
\bibinfo{author}{Storm, P.J.}, \bibinfo{author}{Mandjes, M.},
  \bibinfo{author}{van Arem, B.}, \bibinfo{year}{2022}.
\newblock \bibinfo{title}{Efficient evaluation of stochastic traffic flow
  models using gaussian process approximation}.
\newblock \bibinfo{journal}{Transportation research part B: methodological}
  \bibinfo{volume}{164}, \bibinfo{pages}{126--144}.
%Type = Article
\bibitem[{Tak et~al.(2016)Tak, Woo and Yeo}]{tak2016data}
\bibinfo{author}{Tak, S.}, \bibinfo{author}{Woo, S.}, \bibinfo{author}{Yeo,
  H.}, \bibinfo{year}{2016}.
\newblock \bibinfo{title}{Data-driven imputation method for traffic data in
  sectional units of road links}.
\newblock \bibinfo{journal}{IEEE Transactions on Intelligent Transportation
  Systems} \bibinfo{volume}{17}, \bibinfo{pages}{1762--1771}.
%Type = Article
\bibitem[{Thodi et~al.(2022)Thodi, Khan, Jabari and
  Men{\'e}ndez}]{thodi2022incorporating}
\bibinfo{author}{Thodi, B.T.}, \bibinfo{author}{Khan, Z.S.},
  \bibinfo{author}{Jabari, S.E.}, \bibinfo{author}{Men{\'e}ndez, M.},
  \bibinfo{year}{2022}.
\newblock \bibinfo{title}{Incorporating kinematic wave theory into a deep
  learning method for high-resolution traffic speed estimation}.
\newblock \bibinfo{journal}{IEEE Transactions on Intelligent Transportation
  Systems} .
%Type = Inproceedings
\bibitem[{Titsias(2009)}]{titsias2009variational}
\bibinfo{author}{Titsias, M.}, \bibinfo{year}{2009}.
\newblock \bibinfo{title}{Variational learning of inducing variables in sparse
  gaussian processes}, in: \bibinfo{booktitle}{Artificial intelligence and
  statistics}, \bibinfo{organization}{PMLR}. pp. \bibinfo{pages}{567--574}.
%Type = Article
\bibitem[{Treiber and Helbing(2002)}]{treiber2002reconstructing}
\bibinfo{author}{Treiber, M.}, \bibinfo{author}{Helbing, D.},
  \bibinfo{year}{2002}.
\newblock \bibinfo{title}{Reconstructing the spatio-temporal traffic dynamics
  from stationary detector data}.
\newblock \bibinfo{journal}{Cooperative Transportation Dynamics}
  \bibinfo{volume}{1}, \bibinfo{pages}{3--1}.
%Type = Article
\bibitem[{Treiber et~al.(2000)Treiber, Hennecke and
  Helbing}]{treiber2000congested}
\bibinfo{author}{Treiber, M.}, \bibinfo{author}{Hennecke, A.},
  \bibinfo{author}{Helbing, D.}, \bibinfo{year}{2000}.
\newblock \bibinfo{title}{Congested traffic states in empirical observations
  and microscopic simulations}.
\newblock \bibinfo{journal}{Physical review E} \bibinfo{volume}{62},
  \bibinfo{pages}{1805}.
%Type = Article
\bibitem[{Treiber et~al.(2011)Treiber, Kesting and
  Wilson}]{treiber2011reconstructing}
\bibinfo{author}{Treiber, M.}, \bibinfo{author}{Kesting, A.},
  \bibinfo{author}{Wilson, R.E.}, \bibinfo{year}{2011}.
\newblock \bibinfo{title}{Reconstructing the traffic state by fusion of
  heterogeneous data}.
\newblock \bibinfo{journal}{Computer-Aided Civil and Infrastructure
  Engineering} \bibinfo{volume}{26}, \bibinfo{pages}{408--419}.
%Type = Article
\bibitem[{Usama et~al.(2022)Usama, Ma, Hart and Wojcik}]{usama2022physics}
\bibinfo{author}{Usama, M.}, \bibinfo{author}{Ma, R.}, \bibinfo{author}{Hart,
  J.}, \bibinfo{author}{Wojcik, M.}, \bibinfo{year}{2022}.
\newblock \bibinfo{title}{Physics-informed neural networks (pinns)-based
  traffic state estimation: An application to traffic network}.
\newblock \bibinfo{journal}{Algorithms} \bibinfo{volume}{15},
  \bibinfo{pages}{447}.
%Type = Article
\bibitem[{Van~Hinsbergen et~al.(2011)Van~Hinsbergen, Schreiter, Zuurbier,
  Van~Lint and Van~Zuylen}]{van2011localized}
\bibinfo{author}{Van~Hinsbergen, C.P.}, \bibinfo{author}{Schreiter, T.},
  \bibinfo{author}{Zuurbier, F.S.}, \bibinfo{author}{Van~Lint, J.},
  \bibinfo{author}{Van~Zuylen, H.J.}, \bibinfo{year}{2011}.
\newblock \bibinfo{title}{Localized extended kalman filter for scalable
  real-time traffic state estimation}.
\newblock \bibinfo{journal}{IEEE transactions on intelligent transportation
  systems} \bibinfo{volume}{13}, \bibinfo{pages}{385--394}.
%Type = Article
\bibitem[{Vishnoi et~al.(2022)Vishnoi, Nugroho, Taha and
  Claudel}]{vishnoi2022traffic}
\bibinfo{author}{Vishnoi, S.C.}, \bibinfo{author}{Nugroho, S.A.},
  \bibinfo{author}{Taha, A.F.}, \bibinfo{author}{Claudel, C.G.},
  \bibinfo{year}{2022}.
\newblock \bibinfo{title}{Traffic state estimation for connected vehicles using
  the second-order aw-rascle-zhang traffic model}.
\newblock \bibinfo{journal}{arXiv preprint arXiv:2209.02848} .
%Type = Book
\bibitem[{Wackernagel(2003)}]{wackernagel2003multivariate}
\bibinfo{author}{Wackernagel, H.}, \bibinfo{year}{2003}.
\newblock \bibinfo{title}{Multivariate geostatistics: an introduction with
  applications}.
\newblock \bibinfo{publisher}{Springer Science \& Business Media}.
%Type = Article
\bibitem[{Wang et~al.(2021)Wang, Wu, Zhuang and Sun}]{wang2021low}
\bibinfo{author}{Wang, X.}, \bibinfo{author}{Wu, Y.}, \bibinfo{author}{Zhuang,
  D.}, \bibinfo{author}{Sun, L.}, \bibinfo{year}{2021}.
\newblock \bibinfo{title}{Low-rank hankel tensor completion for traffic speed
  estimation}.
\newblock \bibinfo{journal}{arXiv preprint arXiv:2105.11335} .
%Type = Article
\bibitem[{Wang and Papageorgiou(2005)}]{wang2005real}
\bibinfo{author}{Wang, Y.}, \bibinfo{author}{Papageorgiou, M.},
  \bibinfo{year}{2005}.
\newblock \bibinfo{title}{Real-time freeway traffic state estimation based on
  extended kalman filter: a general approach}.
\newblock \bibinfo{journal}{Transportation Research Part B: Methodological}
  \bibinfo{volume}{39}, \bibinfo{pages}{141--167}.
%Type = Book
\bibitem[{Whitham(2011)}]{whitham2011linear}
\bibinfo{author}{Whitham, G.B.}, \bibinfo{year}{2011}.
\newblock \bibinfo{title}{Linear and nonlinear waves}.
\newblock \bibinfo{publisher}{John Wiley \& Sons}.
%Type = Article
\bibitem[{Work et~al.(2010)Work, Blandin, Tossavainen, Piccoli and
  Bayen}]{work2010traffic}
\bibinfo{author}{Work, D.B.}, \bibinfo{author}{Blandin, S.},
  \bibinfo{author}{Tossavainen, O.P.}, \bibinfo{author}{Piccoli, B.},
  \bibinfo{author}{Bayen, A.M.}, \bibinfo{year}{2010}.
\newblock \bibinfo{title}{A traffic model for velocity data assimilation}.
\newblock \bibinfo{journal}{Applied Mathematics Research eXpress}
  \bibinfo{volume}{2010}, \bibinfo{pages}{1--35}.
%Type = Article
\bibitem[{Xu et~al.(2020)Xu, Wei, Peng, Xuan and Guo}]{xu2020ge}
\bibinfo{author}{Xu, D.}, \bibinfo{author}{Wei, C.}, \bibinfo{author}{Peng,
  P.}, \bibinfo{author}{Xuan, Q.}, \bibinfo{author}{Guo, H.},
  \bibinfo{year}{2020}.
\newblock \bibinfo{title}{Ge-gan: A novel deep learning framework for road
  traffic state estimation}.
\newblock \bibinfo{journal}{Transportation Research Part C: Emerging
  Technologies} \bibinfo{volume}{117}, \bibinfo{pages}{102635}.
%Type = Article
\bibitem[{Yang et~al.(2023)Yang, Ramana and Jabari}]{yang2023generalized}
\bibinfo{author}{Yang, C.}, \bibinfo{author}{Ramana, A.S.V.},
  \bibinfo{author}{Jabari, S.E.}, \bibinfo{year}{2023}.
\newblock \bibinfo{title}{Generalized adaptive smoothing based neural network
  architecture for traffic state estimation}.
\newblock \bibinfo{journal}{IFAC-PapersOnLine} \bibinfo{volume}{56},
  \bibinfo{pages}{3483--3490}.
%Type = Inproceedings
\bibitem[{Yang et~al.(2022)Yang, Thodi and Jabari}]{yang2022generalized}
\bibinfo{author}{Yang, C.}, \bibinfo{author}{Thodi, B.T.},
  \bibinfo{author}{Jabari, S.E.}, \bibinfo{year}{2022}.
\newblock \bibinfo{title}{Generalized adaptive smoothing using matrix
  completion for traffic state estimation}, in: \bibinfo{booktitle}{2022 IEEE
  25th International Conference on Intelligent Transportation Systems (ITSC)},
  \bibinfo{organization}{IEEE}. pp. \bibinfo{pages}{787--792}.
%Type = Article
\bibitem[{Yin et~al.(2012)Yin, Murray-Tuite and Rakha}]{yin2012imputing}
\bibinfo{author}{Yin, W.}, \bibinfo{author}{Murray-Tuite, P.},
  \bibinfo{author}{Rakha, H.}, \bibinfo{year}{2012}.
\newblock \bibinfo{title}{Imputing erroneous data of single-station loop
  detectors for nonincident conditions: Comparison between temporal and spatial
  methods}.
\newblock \bibinfo{journal}{Journal of Intelligent Transportation Systems}
  \bibinfo{volume}{16}, \bibinfo{pages}{159--176}.
%Type = Article
\bibitem[{Yuan et~al.(2014)Yuan, Van~Lint, Van Wageningen-Kessels and
  Hoogendoorn}]{yuan2014network}
\bibinfo{author}{Yuan, Y.}, \bibinfo{author}{Van~Lint, H.},
  \bibinfo{author}{Van Wageningen-Kessels, F.}, \bibinfo{author}{Hoogendoorn,
  S.}, \bibinfo{year}{2014}.
\newblock \bibinfo{title}{Network-wide traffic state estimation using loop
  detector and floating car data}.
\newblock \bibinfo{journal}{Journal of Intelligent Transportation Systems}
  \bibinfo{volume}{18}, \bibinfo{pages}{41--50}.
%Type = Article
\bibitem[{Yuan et~al.(2021)Yuan, Zhang, Yang and Zhe}]{yuan2021macroscopic}
\bibinfo{author}{Yuan, Y.}, \bibinfo{author}{Zhang, Z.}, \bibinfo{author}{Yang,
  X.T.}, \bibinfo{author}{Zhe, S.}, \bibinfo{year}{2021}.
\newblock \bibinfo{title}{Macroscopic traffic flow modeling with physics
  regularized gaussian process: A new insight into machine learning
  applications in transportation}.
\newblock \bibinfo{journal}{Transportation Research Part B: Methodological}
  \bibinfo{volume}{146}, \bibinfo{pages}{88--110}.
%Type = Article
\bibitem[{Zhang and Sun(2024)}]{zhang2024bayesian}
\bibinfo{author}{Zhang, C.}, \bibinfo{author}{Sun, L.}, \bibinfo{year}{2024}.
\newblock \bibinfo{title}{Bayesian calibration of the intelligent driver
  model}.
\newblock \bibinfo{journal}{IEEE Transactions on Intelligent Transportation
  Systems} .
%Type = Article
\bibitem[{Zhang(2002)}]{zhang2002non}
\bibinfo{author}{Zhang, H.M.}, \bibinfo{year}{2002}.
\newblock \bibinfo{title}{A non-equilibrium traffic model devoid of gas-like
  behavior}.
\newblock \bibinfo{journal}{Transportation Research Part B: Methodological}
  \bibinfo{volume}{36}, \bibinfo{pages}{275--290}.
%Type = Article
\bibitem[{Zhong et~al.(2004)Zhong, Lingras and Sharma}]{zhong2004estimation}
\bibinfo{author}{Zhong, M.}, \bibinfo{author}{Lingras, P.},
  \bibinfo{author}{Sharma, S.}, \bibinfo{year}{2004}.
\newblock \bibinfo{title}{Estimation of missing traffic counts using factor,
  genetic, neural, and regression techniques}.
\newblock \bibinfo{journal}{Transportation Research Part C: Emerging
  Technologies} \bibinfo{volume}{12}, \bibinfo{pages}{139--166}.

\end{thebibliography}

\end{document}